\documentclass[runningheads]{llncs}

\usepackage{eccv}

\usepackage{eccvabbrv}

\usepackage{graphicx}
\usepackage{booktabs}

\usepackage{tikz}
\usetikzlibrary{positioning, calc, arrows.meta, backgrounds, fit, shapes.geometric}
\usepackage{multirow}
\usepackage{makecell}
\usepackage{dirtytalk}
\usepackage{mathtools}
\usepackage{pifont}
\newcommand{\cmark}{\ding{51}}
\newcommand{\xmark}{\ding{55}}
\newcommand{\mycomment}[1]{}

\usepackage[accsupp]{axessibility}  

\usepackage{hyperref}
\hypersetup{
  breaklinks=true,
  colorlinks=true,
  citecolor=eccvblue
}

\usepackage{orcidlink}

\begin{document}

\title{Versatile Recompression-Aware Perceptual Image Super-Resolution} 

\titlerunning{Versatile Recompression-Aware Perceptual Image Super-Resolution}




\author{
Mingwei He\inst{1,2}\textsuperscript{*} \and
Tongda Xu\inst{1}\textsuperscript{*} \and
Xingtong Ge\inst{3} \and
Ming Sun\inst{4} \and
Chao Zhou\inst{4} \and
Yan Wang\inst{1}\textsuperscript{\textdagger}
}

\authorrunning{M.~He et al.}

\institute{
AIR, Tsinghua University, Beijing, China
\email{x.tongda@nyu.edu, wangyan@air.tsinghua.edu.cn}
\and
AGUS Tech
\email{mwhe@agus-tech.com}
\and
HKUST, Hong Kong, China
\and
Kuaishou Technology, Beijing, China\\
\textsuperscript{*} Equal contribution. \quad
\textsuperscript{\textdagger} Corresponding author.
}

\maketitle

\begin{center}
  \begin{minipage}[b]{0.55\linewidth}
    \includegraphics[width=\linewidth]{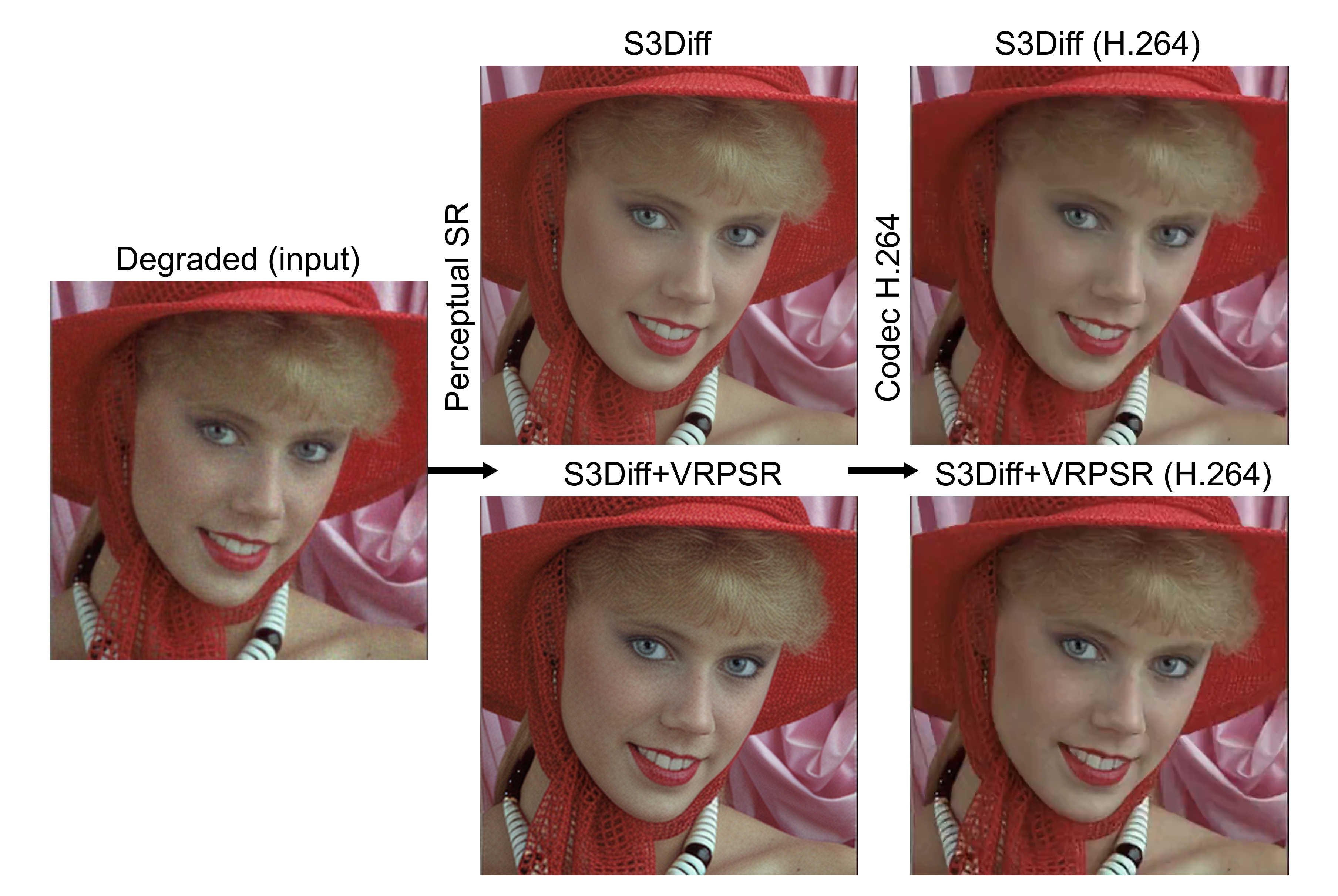}
  \end{minipage}\hspace{0.5em}
  \begin{minipage}[b]{0.42\linewidth}
    \includegraphics[width=\linewidth]{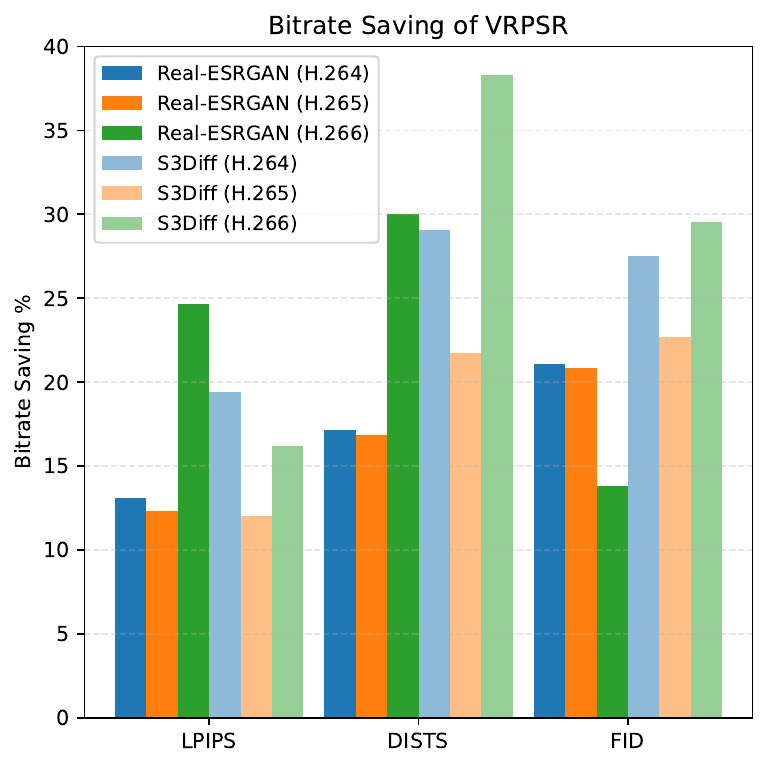}
  \end{minipage}

  {\captionsetup{type=figure}
   \caption{%
     \textit{left:} An example of recompression-aware perceptual super-resolution. The degraded image is first restored and then recompressed. VRPSR optimizes the perceptual quality of the recompressed result. \textit{right:} VRPSR achieves 10\%–40\% bitrate savings when evaluated with super-resolution methods such as Real-ESRGAN \cite{wang2021realesrgan} and S3Diff \cite{2024s3diff}, with codecs including H.264, H.265, and H.266 single-picture mode (intra).}
   \label{fig:teaser}}
\end{center}

\begin{abstract}
Perceptual image super-resolution (SR) methods restore degraded images and produce sharp outputs. In practice, those outputs are usually recompressed for storage and transmission. Ignoring recompression is suboptimal as the downstream codec might add additional artifacts to restored images. However, jointly optimizing SR and recompression is challenging, as the codecs are not differentiable and vary in configuration. In this paper, we present \textbf{Versatile Recompression-Aware Perceptual Super-Resolution (VRPSR)}, which makes existing perceptual SR aware of versatile compression. First, we formulate compression as conditional text-to-image generation and utilize a pre-trained diffusion model to build a generalizable codec simulator. Next, we propose a set of training techniques tailored for perceptual SR, including optimizing the simulator using perceptual targets and adopting slightly compressed images as the training target. Empirically, our VRPSR achieves 10\%–40\% bitrate savings based on Real-ESRGAN and S3Diff under H.264/H.265/H.266 single-picture (intra) compression. Besides, our VRPSR facilitates joint optimization of SR and the post-processing model after recompression. 
\end{abstract}

\begin{figure*}[thb]
    \centering
    \includegraphics[width=\linewidth]{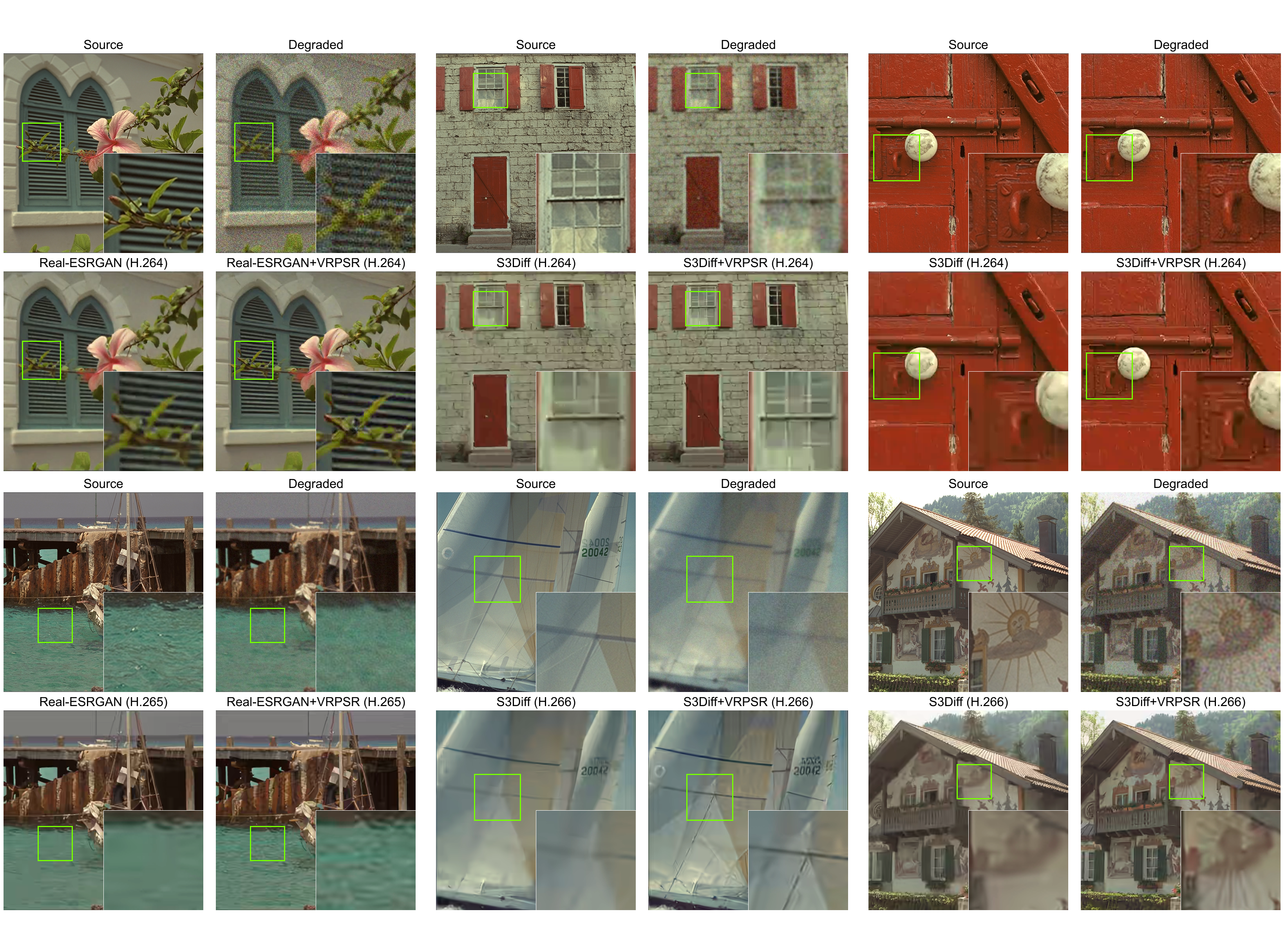}
    \caption{Qualitative results of VRPSR on S3Diff and Real-ESRGAN under H.264, H.265, and H.266 compression across different bitrates. It is shown that after compression, the SR methods with VRPSR produce significantly better visual results at the same bitrate.}
    \label{fig:qual1}
\end{figure*}

\section{Introduction}
\label{sec:intro}

Perceptual image super-resolution (SR) methods have shown great success in image restoration tasks \cite{wang2021realesrgan,wang2024exploiting,2024s3diff}, effectively generating sharp details from blurry inputs. However, a significant practical challenge remains: the restored images often need to be compressed again for storage and distribution. Without being aware of recompression, the fine details generated by SR methods might be turned into artifacts, leading to quality decay and a waste of bitrate.

Unfortunately, directly optimizing SR methods in conjunction with compression is not feasible, as image codecs are non-differentiable and vary substantially. While simple differentiable proxies \cite{xing2021invertible, reich2024differentiable} exist for basic codecs such as JPEG \cite{wallace1991jpeg}, constructing accurate and generalizable differentiable proxies for modern codecs, such as H.265 \cite{sullivan2012overview} and H.266 \cite{bross2021overview}, remains challenging. On the other hand, existing proxy-based differentiable codec simulators \cite{lu2024preprocessing,khan2025perceptual} typically require separate training for each codec type and bitrate setting. A versatile codec proxy that generalizes across diverse conditions remains under-explored.

To address these challenges, we propose Versatile Recompression-Aware Perceptual Super-Resolution (\textbf{VRPSR}) to make existing perceptual image SR aware of recompression. First, we formulate image compression as a conditional text-to-image, which allows us to utilize existing pre-trained diffusion models. Next, we propose a set of training techniques tailored for perceptual SR, such as using a perceptual target to train the simulator and adopting a slightly compressed image as supervision. Empirically, our VRPSR achieves 10\%–40\% bitrate savings on multiple SR algorithms and codecs. Furthermore, our accurate codec simulator enables the joint optimization of SR and post-processing after compression.

Our contributions can be summarized as follows:
\begin{itemize}
\item We propose Versatile Recompression-Aware Perceptual Super-Resolution, the first technique to make perceptual SR algorithms aware of recompression.
\item We formulate image compression as a conditional text-to-image generation and utilize a pre-trained diffusion model for accurate and generalizable codec simulation.
\item We introduce a set of training strategies tailored for perceptual SR, such as using a perceptual target for the simulator and training the SR model on slightly compressed images.
\item Empirically, we demonstrate that VRPSR enhances the performance of multiple perceptual SR methods with a range of codecs for recompression.
\item Furthermore, our accurate simulator enables the joint optimization of the SR model and an additional post-processing module after recompression.
\end{itemize}

\begin{figure}[thb]
    \centering
    \includegraphics[width=0.9\linewidth]{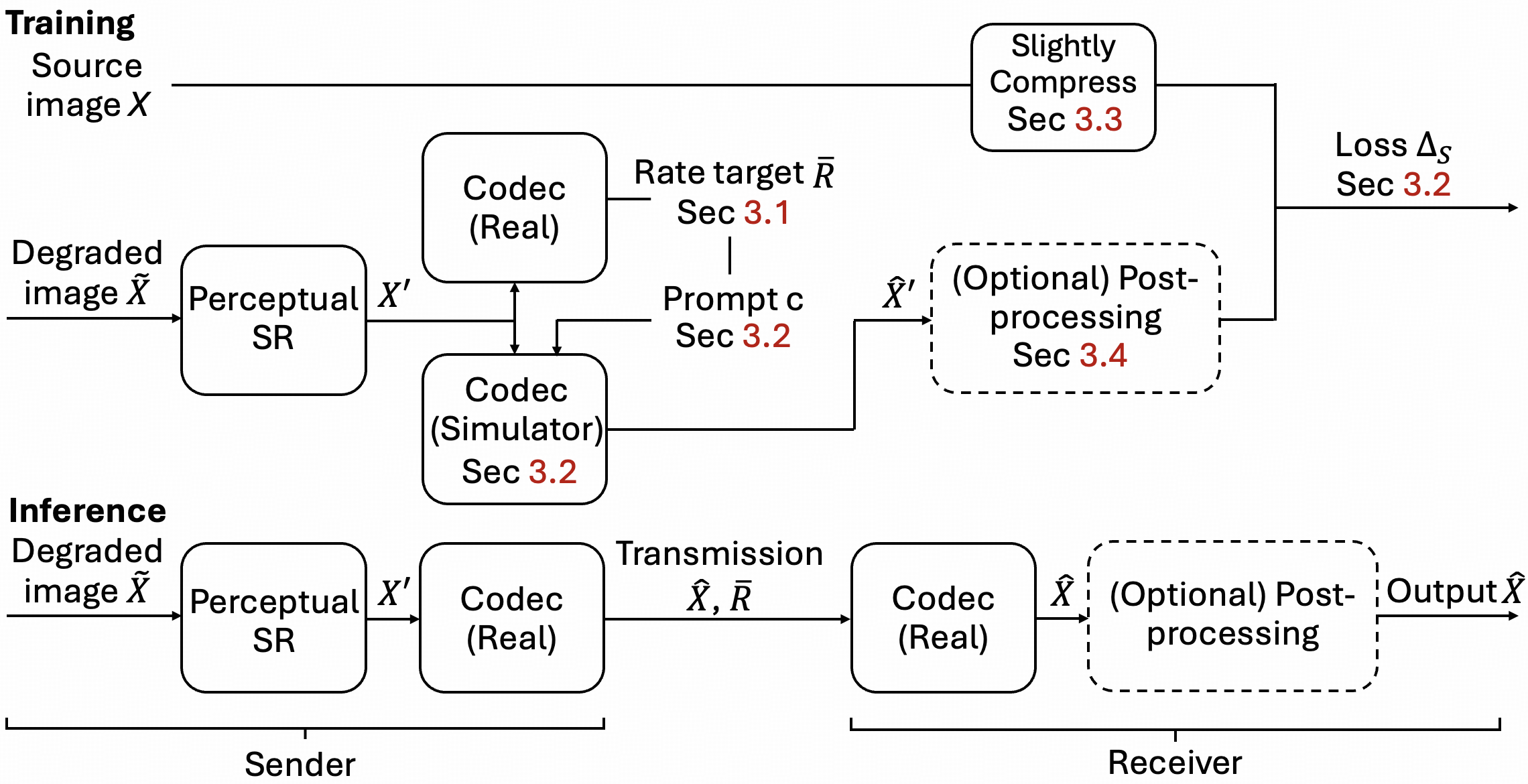}
    \caption{The training and inference pipeline of VRPSR. For training, we adopt a rate target codec simulator and use a slightly compressed image for supervision. For inference, the degraded image $\tilde{X}$ is first restored by a perceptual super-resolution method to obtain $X'$. Then it is compressed by the encoder into $\hat{X}$ with bitrate $\bar{R}$. Then, the compressed image is transmitted from the sender to the receiver. Then we either directly use image $\hat{X}$ as the final output, or optionally include a post-processing module.}
    \label{fig:pipe}
\end{figure}

\section{Related Works}

\subsection{Perceptual Image Super-resolution} Perceptual image SR algorithms aim to achieve blind image restoration in real-world scenarios, where the degradation is unknown and training is typically conducted on synthetically generated data \cite{wang2021realesrgan,zhang2021designing}. Inspired by the perception-distortion trade-off theorem \cite{blau2018perception}, early works in this area \cite{wang2021realesrgan, zhang2021designing} employed conditional Generative Adversarial Networks \cite{goodfellow2020generative} to enhance perceptual quality. More recent approaches \cite{saharia2022image,li2022srdiff,yue2023resshift,wang2024exploiting,wang2024sinsr,2024s3diff,wu2024seesr} adopt diffusion models \cite{ho2020denoising,song2020score} to further improve generative capabilities. However, the task of recompression-aware perceptual SR remains under-explored.

\subsection{Differentiable Image Codec Simulator} Differentiable image codec simulators aim to approximate the gradients of image codecs. These simulators generally fall into two categories: non-learning-based and learning-based. Non-learning-based simulators are typically designed for simple codecs such as JPEG \cite{xing2021invertible,yang2023self,reich2024differentiable,Hu2024StandardCV,guleryuz2024sandwiched} and H.264 \cite{chadha2021deep}, where non-differentiable components, such as quantization and mode decisions, are replaced with differentiable alternatives like the straight-through estimator \cite{bengio2013estimating} and Gumbel softmax \cite{jang2016categorical,maddison2016concrete}. However, for more advanced codecs such as H.265 and H.266, constructing non-learning-based simulators becomes prohibitively complex. Learning-based simulators use neural networks to approximate both the reconstruction and the bitrate produced by a real codec. For instance, author \cite{tian2021self} adopts DenseNet \cite{iandola2014densenet} to approximate the reconstruction of H.265. Subsequent works \cite{lu2024preprocessing, khan2025perceptual} utilize hyperprior-based architectures \cite{balle2018variational} to approximate both rate and reconstruction for different codecs. However, these methods typically require a separate model for each codec type, whereas our VRPSR uses a single model for all codecs.

\section{Versatile Recompression-Aware Perceptual Super-resolution}

\subsection{Problem Formulation}
Let $X$ denote the clean image. We assume that the observed corrupted image $\tilde{X}$ is generated via an unknown kernel $p(\tilde{X}|X)$. After obtaining $\tilde{X}$, the image is compressed by a codec $f(\cdot; q)$ with parameter $q$, resulting in a compressed image $\bar{X}$ with bitrate $\bar{R}$:
\begin{gather}
    \textrm{Corruption: } \tilde{X} \sim p_C(\tilde{X}|X), \notag \\ 
    \textrm{Codec w/o SR: } \bar{X}, \bar{R} = f(\tilde{X}, q).
\end{gather}
To restore $\tilde{X}$ and obtain a higher-quality image, we apply a perceptual SR model $g_{\phi}(.)$, producing a restored image $X'$. In a practical scenario, we need to keep the bitrate of the source image $\tilde{X}$ and the processed image $X'$ similar, as the network bandwidth is fixed. So when recompressing $X'$, the codec parameter $q'$ is adjusted such that the resulting bitrate $\hat{R}$ matches $\bar{R}$. Specifically, we require a rate-preserving compression process:
\begin{gather}
    \textrm{SR: }X' = g_{\phi}(\tilde{X}), \notag \\
    \textrm{Codec w/ SR: } \hat{X}, \hat{R} = f(X', q') \textrm{, select $q'$ s.t. } \hat{R} \approx \bar{R}.\label{eq:c2} 
\end{gather}
The optimization of the recompression-aware perceptual SR model parametrized by $\phi$, is typically formulated as a weighted combination of mean square error (MSE), LPIPS \cite{zhang2018unreasonable}, and adversarial loss \cite{blau2018perception}:
\begin{align}
\phi^* &\leftarrow \arg\min_{\phi}, \Delta_S(\hat{X},X), \textrm{where } 
\Delta_S(\hat{X},X) \notag \\ &= \mathcal{L}_{MSE}(\hat{X},X) +\mathcal{L}_{LPIPS}(\hat{X},X)+\mathcal{L}_{GAN}(\hat{X}).\label{eq:target}
\end{align}
A major challenge arises because the optimization target in Eq.~\ref{eq:target} cannot be directly minimized using gradient descent, as the rate-preserving compression process described in Eq. \ref{eq:c2} is not differentiable, and the codec type and parameter $q$ can vary widely. 

\subsection{Image Codec Simulation as Text to Image}
To optimize Eq.~\ref{eq:target}, we must simulate the rate-preserving image compression process described in Eq.~\ref{eq:c2} using a differentiable function, \textit{i.e.,} computing $d\hat{X}/dX'$. Besides, it is essential to account for the diversity of codecs $f(\cdot; q)$. To address both challenges, we propose formulating codec simulation as a text-to-image problem to utilize a pre-trained diffusion model.

\subsubsection{Rate-Target Codec Simulation} Most existing codec simulators \cite{chadha2021deep,tian2021self,said2022differentiable,lu2024preprocessing,khan2025perceptual} directly approximate the gradient of the codec $f(\cdot; q)$ by training a neural network $f_{\theta}(.,.)$, parameterized by $\theta$, to mimic the behavior of the actual codec for a given encoding parameter $q$. Both the bitrate $\hat{R}$ and the reconstruction $\hat{X}$ need to be learned. However, they cannot directly simulate the rate-preserving compression process in Eq.~\ref{eq:c2} as the selection of $q'$ remains non-differentiable. On the other hand, we propose to directly approximate the output of the rate-preserving encoding process, with the bitrate as input. Formally, given a distortion metric $\Delta_C(.,.)$, the optimization target of our rate-target codec simulator is defined as:
\begin{gather}
    \hat{X}' = f_{\theta}(X',\bar{R}), \notag \\
    \theta^* \leftarrow \arg\min_{\theta} \Delta_C(\hat{X}',\hat{X}) \textrm{, where } \hat{X} \textrm{ is from Eq.~\ref{eq:c2}}. \label{eq:rt}
\end{gather}

\begin{table}[thb]
\caption{Quantitative results on the Kodak dataset. Our VRPSR achieves 10\%–40\% bitrate savings on the perceptual metric.}
\label{tab:quant2}
\centering
\resizebox{\linewidth}{!}{
\begin{tabular}{@{}lccccccccccccccc@{}}
\toprule
 & \multicolumn{5}{c}{H.264 (x264)} & \multicolumn{5}{c}{H.265 (x265)} & \multicolumn{5}{c}{H.266 (vvenc)} \\ \cmidrule(lr){2-6} \cmidrule(lr){7-11} \cmidrule(lr){12-16}
 & bpp & PSNR & LPIPS & DISTS & FID & bpp & PSNR & LPIPS & DISTS & FID & bpp & PSNR & LPIPS & DISTS & FID \\ \midrule
\multirow{4}{*}{Real-ESRGAN}
 & 0.11 & 26.01 & 0.467 & 0.277 & 146.63 & 0.15 & 26.70 & 0.471 & 0.254 & 148.83 & 0.14 & 27.38 & 0.425 & 0.227 & 99.54\\
 & 0.16 & 26.85 & 0.403 & 0.242 & 102.90 & 0.21 & 27.31 & 0.418 & 0.225 & 100.13 & 0.18 & 27.59 & 0.403 & 0.214 & 80.57 \\
 & 0.25 & 27.30 & 0.365 & 0.220 & 82.59  & 0.28 & 27.62 & 0.386 & 0.208 & 78.77  & 0.23 & 27.72 & 0.387 & 0.205 & 71.64 \\ 
 & 0.44 & 27.78 & 0.319 & 0.189 & 62.31  & 0.40 & 27.87 & 0.354 & 0.190 & 63.92  & 0.37 & 27.90 & 0.357 & 0.189 & 61.76 \\
 \midrule
\multirow{5}{*}{\makecell[l]{Real-ESRGAN\\+VRPSR (Ours)}}
 & 0.11 & 25.79 & 0.450 & 0.264 & 137.07 & 0.15 & 26.22 & 0.455 & 0.238 & 132.42 & 0.14 & 27.13 & 0.409 & 0.213 & 86.62 \\
 & 0.16 & 26.64 & 0.388 & 0.230 & 90.19  & 0.21 & 26.99 & 0.400 & 0.212 & 85.40  & 0.18 & 27.39 & 0.385 & 0.202 & 76.30 \\
 & 0.25 & 27.15 & 0.348 & 0.206 & 77.23  & 0.28 & 27.39 & 0.368 & 0.196 & 74.68  & 0.23 & 27.57 & 0.368 & 0.193 & 67.40 \\
 & 0.44 & 27.64 & 0.296 & 0.179 & 57.27  & 0.40 & 27.71 & 0.327 & 0.179 & 57.98  & 0.37 & 27.77 & 0.327 & 0.177 & 58.91 \\
\cmidrule(lr){2-6} \cmidrule(lr){7-11} \cmidrule(lr){12-16}
& BDBR & 17.43\% & -13.07\% & -17.13\% & -21.07\% & BDBR & 26.63\% & -12.31\% & -16.86\% & -20.83\% & BDBR & 33.44\% & -24.68\% & -30.01\% & -13.79\% \\ 
\midrule
\multirow{4}{*}{S3Diff}  
 & 0.09 & 25.39 & 0.501 & 0.295 & 160.73 & 0.13 & 25.95 & 0.495 & 0.269 & 156.18 & 0.11 & 26.50 & 0.445 & 0.242 & 119.55\\
 & 0.13 & 26.05 & 0.436 & 0.259 & 118.89 & 0.17 & 26.43 & 0.436 & 0.236 & 118.03 & 0.14 & 26.64 & 0.413 & 0.222 & 93.11 \\
 & 0.16 & 26.31 & 0.402 & 0.241 & 95.29  & 0.21 & 26.59 & 0.409 & 0.221 & 96.47  & 0.18 & 26.72 & 0.384 & 0.209 & 79.59 \\
 & 0.28 & 26.61 & 0.343 & 0.205 & 66.78  & 0.30 & 26.76 & 0.353 & 0.194 & 69.61  & 0.23 & 26.78 & 0.361 & 0.194 & 66.47 \\
 \midrule
\multirow{5}{*}{\makecell[l]{S3Diff\\+VRPSR (Ours)}} 
 & 0.09 & 24.95 & 0.463 & 0.265 & 139.43 & 0.13 & 25.25 & 0.474 & 0.245 & 129.69 & 0.11 & 25.98 & 0.422 & 0.211 & 86.43 \\
 & 0.13 & 25.66 & 0.404 & 0.232 & 89.52  & 0.17 & 25.99 & 0.413 & 0.215 & 88.83  & 0.14 & 26.36 & 0.390 & 0.196 & 74.11 \\
 & 0.16 & 26.02 & 0.375 & 0.217 & 78.91  & 0.21 & 26.26 & 0.385 & 0.200 & 75.51  & 0.18 & 26.56 & 0.367 & 0.186 & 63.58 \\
 & 0.28 & 26.49 & 0.323 & 0.186 & 58.53  & 0.30 & 26.64 & 0.342 & 0.180 & 56.67  & 0.23 & 26.68 & 0.346 & 0.175 & 54.25 \\
 \cmidrule(lr){2-6} \cmidrule(lr){7-11} \cmidrule(lr){12-16}
& BDBR & 27.05\% & -19.40\% & -29.05\% & -27.50\% & BDBR & 35.24\% & -12.02\% & -21.76\% & -22.70\% & BDBR & 53.67\% & -16.21\% & -38.31\% & -29.54\% \\
\bottomrule
\end{tabular}
}
\end{table}

\begin{figure*}[thb]
    \centering
    \includegraphics[width=\linewidth]{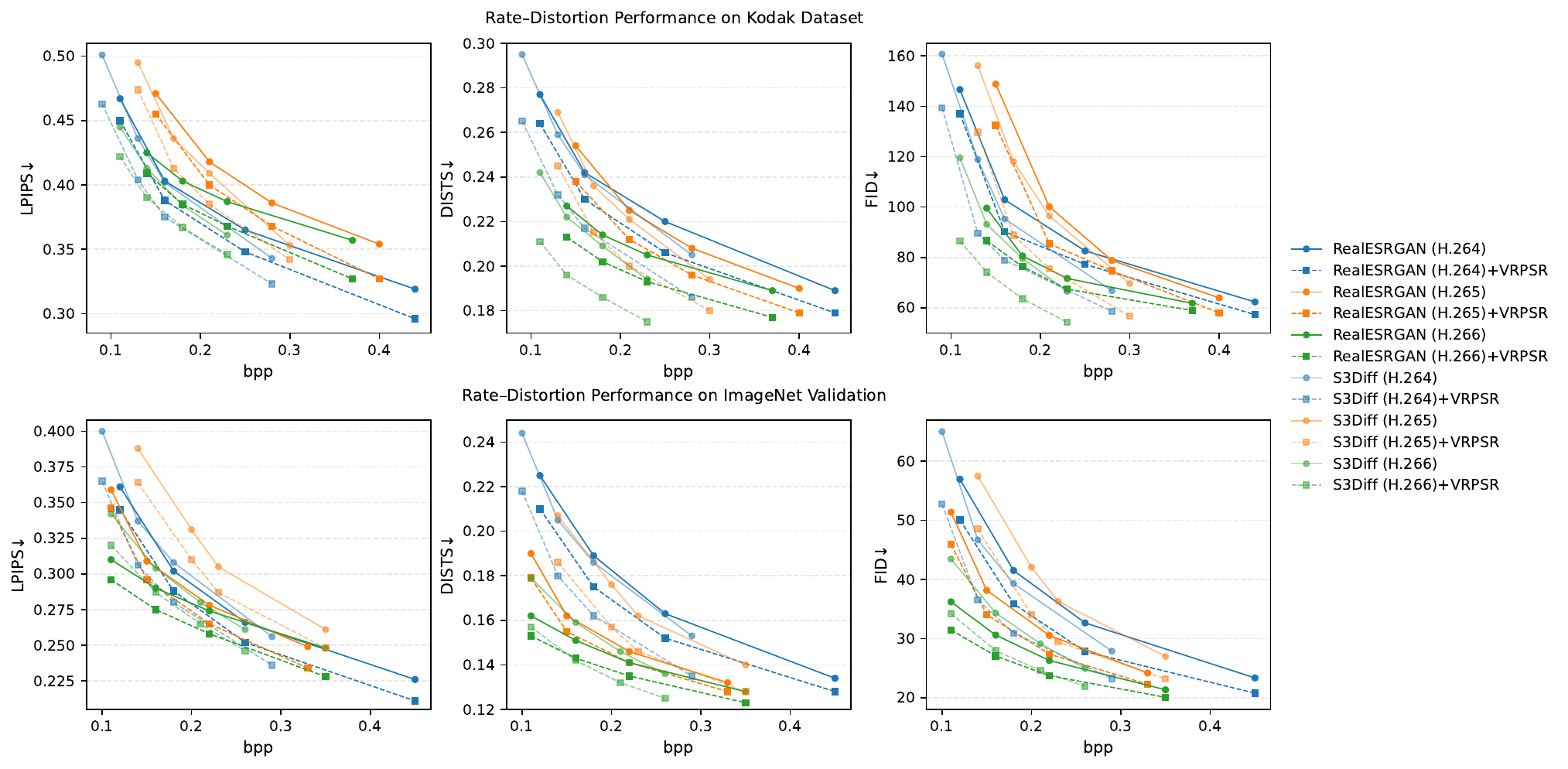}
    \caption{Rate-distortion performance on Kodak and ImageNet Validation dataset. Our VRPSR boosts the performance of RealESRGAN and S3Diff with multiple downstream codecs such as H.264, H.265, and H.266.}
    \label{fig:rd_curves}
\end{figure*}
\subsubsection{Codec Simulation as Text-to-Image Generation} To enhance the versatility of our simulator $f_{\theta}(.)$ across various codec types, we propose to formulate the codec simulation as a conditional text-to-image task. For example, if the codec $f(\cdot; q)$ is H.264 (x264) with a target bitrate $0.3$ bits per pixel (bpp), the corresponding text prompt would be:
\begin{gather}
    c = \texttt{A \{libx264\} \{0.3\} bpp compressed image.} \notag 
\end{gather}
Under this formulation, the rate-target codec simulator in Eq.~\ref{eq:rt} becomes a conditional text-to-image generation:
\begin{gather}
    \hat{X}' = f_{\theta}(X', c). \label{eq:uc}
\end{gather}
This formulation allows the simulator in Eq.~\ref{eq:uc} to leverage powerful pre-trained text-to-image models, such as Stable Diffusion \cite{rombach2022high}, with additional image condition inputs $X'$ as supported by models like ControlNet \cite{zhang2023adding}. In practice, we implement $f_\theta(.,.)$ with S3Diff \cite{2024s3diff}, which is based on  distilled Stable Diffusion \cite{sauer2024adversarial}.

\subsubsection{Additional Codec Condition Embedding}
\label{sec:codec-embed}
In addition to including codec parameters in the prompt $c$, we inject them into SR backbones and the simulator via position embedding. More specifically, we treat the codec type as a one-hot vector, concatenate it with the target bpp, and transform them into an embedding vector similar to the temporal embedding of Stable Diffusion \cite{rombach2022high}. For the Real-ESRGAN backbone, we feed the embedding vector into residual blocks using AdaIN modulation \cite{Huang2017ArbitraryST}. For the S3Diff backbone, we add the embedding vector to the time embedding (see Appendix~\ref{app:arch} for details). Empirically, we find that the additional embedding improves the performance of both the simulator and the final SR model.

\subsection{Training Techniques for Perceptual SR}
Empirically, we find that following previous simulators training \cite{tian2021self,khan2025perceptual} that are designed for MSE-oriented SR is suboptimal for perceptual SR. Therefore, we propose several techniques tailored for training recompression-aware perceptual SR.

\subsubsection{Perceptual Training Target for the Simulator}
Most existing approaches train codec proxies using MSE between the simulator output and the ground-truth codec output, even when the simulator is intended for perceptual applications \cite{tian2021self,khan2025perceptual}. In contrast, we demonstrate that training the simulator with the perceptual target yields better results. Accordingly, we align the optimization target of the simulator in Eq.~\ref{eq:rt} with the target in SR training:
\begin{align}    
    \Delta_C(\hat{X}',\hat{X}) = \Delta_S(\hat{X}',\hat{X}). \label{eq:ptgt}
\end{align}
As shown in Table~\ref{tab:simu}, optimizing with perceptual loss has minimal impact on PSNR but leads to significant improvements in perceptual quality metrics.

\subsubsection{Slightly Compressed Image for Supervision} By default, perceptual SR training is supervised against the clean image $X$. However, we empirically find that this leads to unstable training and suboptimal performance, likely because the clean image is too far from the reconstruction space of a practical codec. Therefore, we propose to train the compression-aware SR model using a slightly compressed image. Specifically, we process $X$ with our simulator using a configuration $c'$ that corresponds to a slightly stronger compression setting (e.g., $q-10$):

\begin{align}
    X_{s}= f_{\theta}(X,c').
\end{align}

We further analyze the sensitivity to the supervision compression strength in Appendix~\ref{app:q_sensitivity}.

\subsubsection{Two Stage Optimization} We adopt a two-stage optimization scheme for training the codec simulator and recompression-aware SR. First, the codec simulator is trained independently using the compressed outputs of ground-truth images $X$. Next, the SR parameter $\phi(.)$ is updated iteratively with the simulator parameter $\theta$ frozen. During SR updates, we use the output of SR $X'$, rather than the ground-truth images, as input. Unlike prior works \cite{khan2025perceptual}, we find that jointly updating the simulator and SR model is harmful to performance and stick to separate optimization.

\subsubsection{No Straight-Through Estimator}
Once the simulator output $\hat{X}' = f_{\theta}(X',c)$ is obtained, the most straightforward strategy is to directly use $\hat{X}'$ in place of the true codec output $\hat{X}$ in Eq.~\ref{eq:target}. In contrast, many prior works \cite{lu2024preprocessing,khan2025perceptual} adopt the straight-through estimator (STE) \cite{bengio2013estimating}, claiming improved performance. STE is a training technique that replaces the simulator output $\hat{X}'$ with true codec output $\hat{X}$ during the forward pass, while preserving the gradient of the simulator during backpropagation. This can be implemented using a stop-gradient operator from the autograd package $\textrm{sg}(.)$:
\begin{gather}
\textrm{no STE: } \hat{X} = \hat{X}' \notag, \\
\textrm{STE: }\hat{X}_{\mathrm{STE}} = \hat{X} + \hat{X}' - \textrm{sg}(\hat{X}').
\end{gather}
We conduct a comprehensive study on the impact of using STE. Our findings reveal that using the simulator output directly without STE yields better performance.

\subsection{Optional Joint Post-Processing}\label{sec:sandwich}
Despite VRPSR improving the perceptual SR's performance with recompression, the overall performance of VRPSR is still bounded by the final codec. When the bitrate target is low, extremely heavy recompression still produces images with strong artifacts (see Appendix~\ref{app:quant} for details). To tackle this challenge, we further propose an optional post-processing module after the recompression as shown in Figure~\ref{fig:pipe}. More specifically, we propose an additional post-processing network after the codec output $\hat{X}$, with prompt $c$ as condition:
\begin{gather}
    \hat{X}_{post} = h_{\phi}(\hat{X},c).
\end{gather}
With a differentiable codec simulator, the perceptual SR and post-processing module can be jointly optimized. Empirically, we show that such joint optimization leads to favourable results both quantitatively and qualitatively.

\begin{figure*}[thb]
    \centering 
    \includegraphics[width=1.0\linewidth]{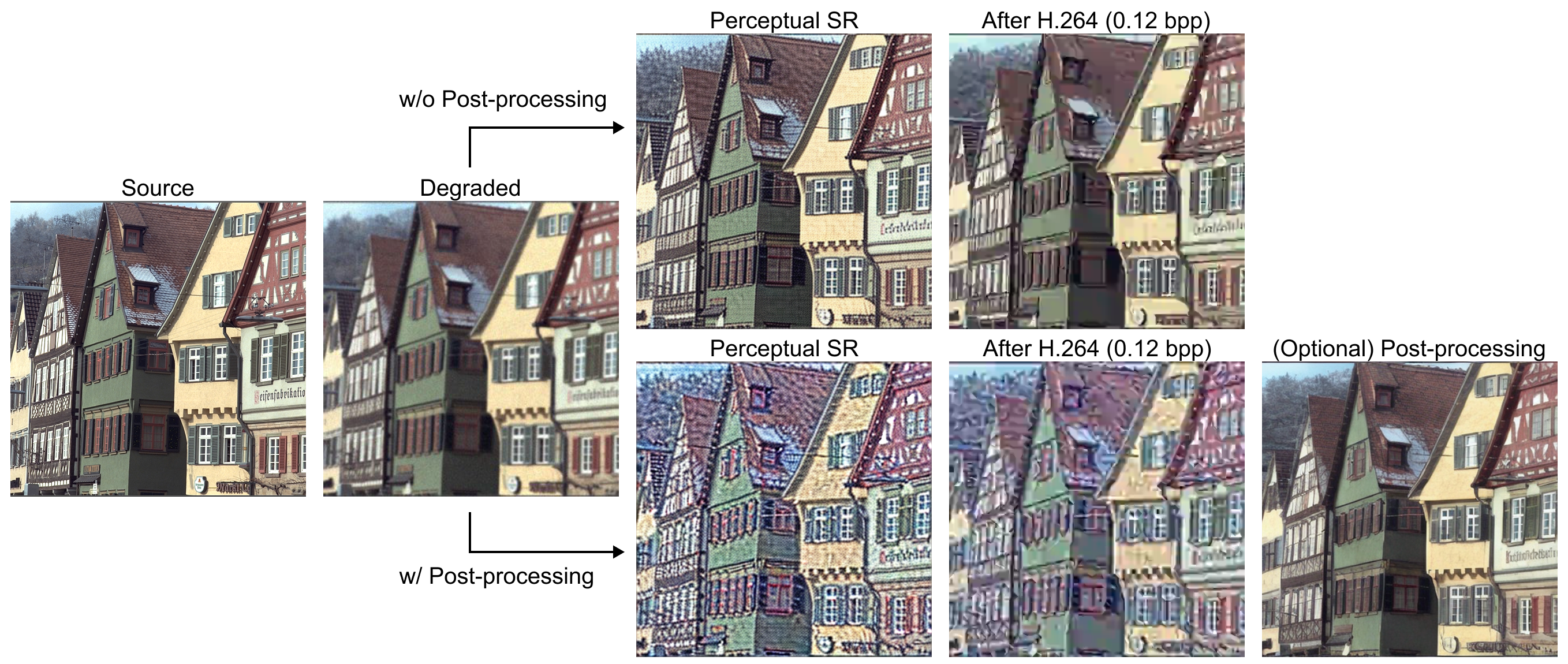}
    \caption{Qualitative results of VRPSR on S3Diff under H.264 recompression. The final output with the post-processing module effectively removes codec artifacts and restores perceptual details.}
    \label{fig:post_pipe_pic}
\end{figure*}

\section{Experimental Results}

\subsection{Experimental Setup} For SR baselines, we select \textbf{Real-ESRGAN} \cite{wang2021realesrgan} and \textbf{S3Diff} \cite{2024s3diff}, which are two representative perceptual SR methods based on GANs and diffusion, respectively. All models are trained on the \textbf{ImageNet} \cite{Deng2009ImageNetAL} training split. For evaluation, we use the $\textbf{Kodak}$ \cite{kodak} and the $\textbf{ImageNet}$ validation split. All images are resized and center-cropped to $512\times 512$. We evaluate performance using the standard metrics for perceptual image SR: Peak Signal-to-Noise Ratio (\textbf{PSNR}), Learned Perceptual Image Patch Similarity (\textbf{LPIPS}) \cite{Zhang2018TheUE}, Deep Image Structure and Texture Similarity (\textbf{DISTS}) \cite{ding2020image}, and Fréchet Inception Distance (\textbf{FID}) \cite{Heusel2017GANsTB}. We also compute Structural Similarity (\textbf{SSIM}) \cite{wang2004image} and report it in the supplementary material (Appendix~\ref{app:quant}). To assess performance across varying bitrates, we additionally compute the Bjontegaard Bitrate (\textbf{BDBR}), a widely used metric for measuring average bitrate savings \cite{bjontegaard2001calculation}. For simulator comparison, we include \textbf{JPEG} based simulator \cite{guleryuz2024sandwiched,Hu2024StandardCV} and \textbf{Hyper-prior} based simulators \cite{lu2024preprocessing,khan2025perceptual}, which are among the most commonly adopted in recent literature. For the downstream codec, we select three widely used compression standards of increasing complexity: \textbf{H.264 (x264)}, \textbf{H.265 (x265)}, and \textbf{H.266 (vvenc)}. We exclude JPEG from our primary evaluation due to its simplicity and the availability of specialized simulators \cite{reich2024differentiable}. See Appendix~\ref{app:impl} for more details.

\subsection{Results of VRPSR} In Table~\ref{tab:quant2}, Figure~\ref{fig:qual1}, and Figure~\ref{fig:rd_curves}, we show the effectiveness of VRPSR applied to both Real-ESRGAN and S3Diff on the Kodak and ImageNet datasets. Compared to baseline models trained without recompression-aware strategies, VRPSR achieves substantial bitrate savings—ranging from $10\%-40\%$ —when evaluated using perceptual metrics such as LPIPS, DISTS, and FID. These improvements are consistent across different codec types (H.264, H.265, and H.266) and a wide range of bitrates (from $0.09$ to $0.45$ bpp). Additional quantitative results on the ImageNet validation set are provided in the supplementary material (Table~\ref{tab:quant1}). As shown in Figure~\ref{fig:qual1}, the visual improvements introduced by VRPSR are also significant and clearly perceptible.

\begin{table}[htb]
    \centering
    \includegraphics[width=0.9\linewidth]{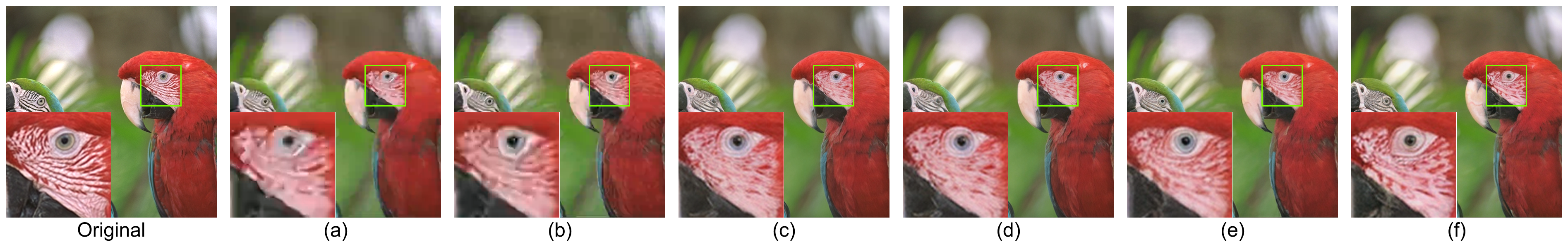}
    \captionof{figure}{Qualitative results of optional post-processing. See definition of (a)-(f) in Table~\ref{tab:sandwich}.}
    \label{fig:sandwich-qual}
    \captionof{table}{Quantitative results of optional post-processing. Jointly training SR and post-processing brings the best result. \textbf{Bold}: Best.}
    \vspace{1em}
    \label{tab:sandwich}
    \centering
    \resizebox{1.0\linewidth}{!}{
    \begin{tabular}{@{}cccccccccc@{}}
    \toprule
    \multirow{2}{*}{ID} & \multirow{2}{*}{Perceptual SR} & \multirow{2}{*}{VRPSR} & \multirow{2}{*}{Post-processing} & \multirow{2}{*}{Optimization} & \multicolumn{5}{c}{S3Diff + H.264} \\ \cmidrule(l){6-10} 
     & &  &  &  & bpp & PSNR & LPIPS & DISTS & FID \\ \midrule
    (a) & \cmark & \xmark & \xmark & SR only      & 0.16 & \textbf{26.31} & 0.402 & 0.241 & 95.29 \\
    (b) & \cmark & \cmark & \xmark & SR only      & 0.16 & 26.02 & 0.375 & 0.217 & 78.91 \\
    (c) & \xmark & \xmark & \cmark & Post only    & 0.16 & 26.09 & 0.275 & 0.157 & 52.08 \\
    (d) & \cmark & \xmark & \cmark & SR then Post & 0.16 & 26.01 & 0.286 & 0.166 & 53.36 \\
    (e) & \cmark & \cmark & \cmark & SR then Post & 0.16 & 25.81 & 0.274 & 0.157 & 52.81 \\
    (f) & \cmark & \cmark & \cmark & Joint        & 0.16 & 25.76 & \textbf{0.241} & \textbf{0.144} & \textbf{49.31} \\ \bottomrule
        \end{tabular}
    }
\end{table}

\subsection{Results of VRPSR with Post-processing} Despite VRPSR successfully improving perceptual SR when facing recompression, its performance is still bounded by the codec, especially in the low bitrate regime. In that case, our optional post-processing module can further boost the performance of VRPSR. As shown in Table~\ref{tab:sandwich}, we compare several different ways to train post-processing. In Figure~\ref{fig:sandwich-qual} and Table~\ref{tab:sandwich}, we first confirm that the perceptual SR with VRPSR is much better than the perceptual SR without recompression awareness (Table~\ref{tab:sandwich} (a) vs (b)). Next, we show that training post-processing alone is better than SR alone (Table~\ref{tab:sandwich} (c) vs (b)), and training post-processing on top of perceptual SR is better than post-processing alone (Table~\ref{tab:sandwich} (d) (e) vs (c)). Finally, we show that joint training of post-processing and perceptual SR with a gradient estimated by VRPSR is the best (Table~\ref{tab:sandwich} (f) vs (d)(e)).

\begin{table*}[thb]
\caption{Ablation study on the design components of our proposed simulator. Results are reported on the Kodak set.}
\label{tab:simu}
\centering
\resizebox{\linewidth}{!}{
\begin{tabular}{@{}ccccccccccccccc@{}}
\toprule
\multirow{2}{*}{ID} & \multirow{2}{*}{Pre-train} & \multirow{2}{*}{Prompt} & \multirow{2}{*}{Embed} & \multirow{2}{*}{Output} & \multirow{2}{*}{Target} & \multicolumn{4}{c}{Simulator} & \multicolumn{5}{c}{ S3Diff+H.264} \\ \cmidrule(lr){7-10} \cmidrule(lr){11-15}
&  (Sec 3.2.2) & (Sec 3.2.2) & (Sec 3.2.3) & (Sec 3.2.1) & (Sec 3.3.1) & PSNR & LPIPS & DISTS & FID & bpp & PSNR & LPIPS & DISTS & FID \\
\midrule
(a) & \xmark & \cmark                & \cmark                & $\hat{X}'\textrm{, keep } \hat{R}'=\bar{R}$ & Perceptual & 26.35          & 0.374          & 0.213          & 117.09         & 0.10 & 25.58          & 0.498          & 0.285          & 151.12 \\
(b) & \cmark & \cmark(no $\hat{R}'$) & \cmark(no $\hat{R}'$) & $\hat{X}',\hat{R}'$                         & Perceptual & 28.67          & 0.168          & 0.116          & 61.22          & 0.10 & \textbf{25.66} & 0.479          & 0.282          & 150.41 \\
(c) & \cmark & \xmark                & \xmark                & $\hat{X}'\textrm{, keep } \hat{R}'=\bar{R}$ & Perceptual & 28.85          & 0.209          & 0.164          & 89.98          & 0.10 & 25.61          & 0.482          & 0.284          & 154.30 \\
(d) & \cmark & \cmark                & \xmark                & $\hat{X}'\textrm{, keep } \hat{R}'=\bar{R}$ & Perceptual & 28.73          & 0.161          & 0.103          & 54.33          & 0.10 & 25.15          & 0.460          & 0.269          & 143.62 \\
(e) & \cmark & \cmark                & \cmark                & $\hat{X}'\textrm{, keep } \hat{R}'=\bar{R}$ & MSE        & \textbf{30.93} & 0.192          & 0.167          & 92.31          & 0.10 & 25.42          & 0.461          & 0.274          & 165.82 \\
(f) & \cmark & \cmark                & \cmark                & $\hat{X}'\textrm{, keep } \hat{R}'=\bar{R}$ & Perceptual & 29.43          & \textbf{0.157} & \textbf{0.101} & \textbf{50.13} & 0.10 & 24.95          & \textbf{0.459} & \textbf{0.264} & \textbf{130.99} \\
\bottomrule
\end{tabular}
}
\end{table*}

\begin{table}[htb]
\caption{Comparison with other simulators.}
\label{tab:simu2}
\centering
\resizebox{0.45\linewidth}{!}{
\begin{tabular}{@{}lccccc@{}}
\toprule
\multirow{2}{*}{Simulator} & \multicolumn{5}{c}{S3Diff (H.264)} \\ \cmidrule(lr){2-6}
& bpp & PSNR & LPIPS & DISTS & FID \\ \midrule
JPEG         & 0.13 & 25.98 & 0.436 & 0.255 & 125.92 \\
Hyperprior   & 0.13 & \textbf{26.22} & 0.445 & 0.257 & 117.49 \\
VRPSR (Ours) & 0.13 & 25.66 & \textbf{0.404} & \textbf{0.232} & \textbf{89.52} \\
 \bottomrule
\end{tabular}
}
\end{table}

\begin{figure}[htb]
    \centering
    \includegraphics[width=\linewidth]{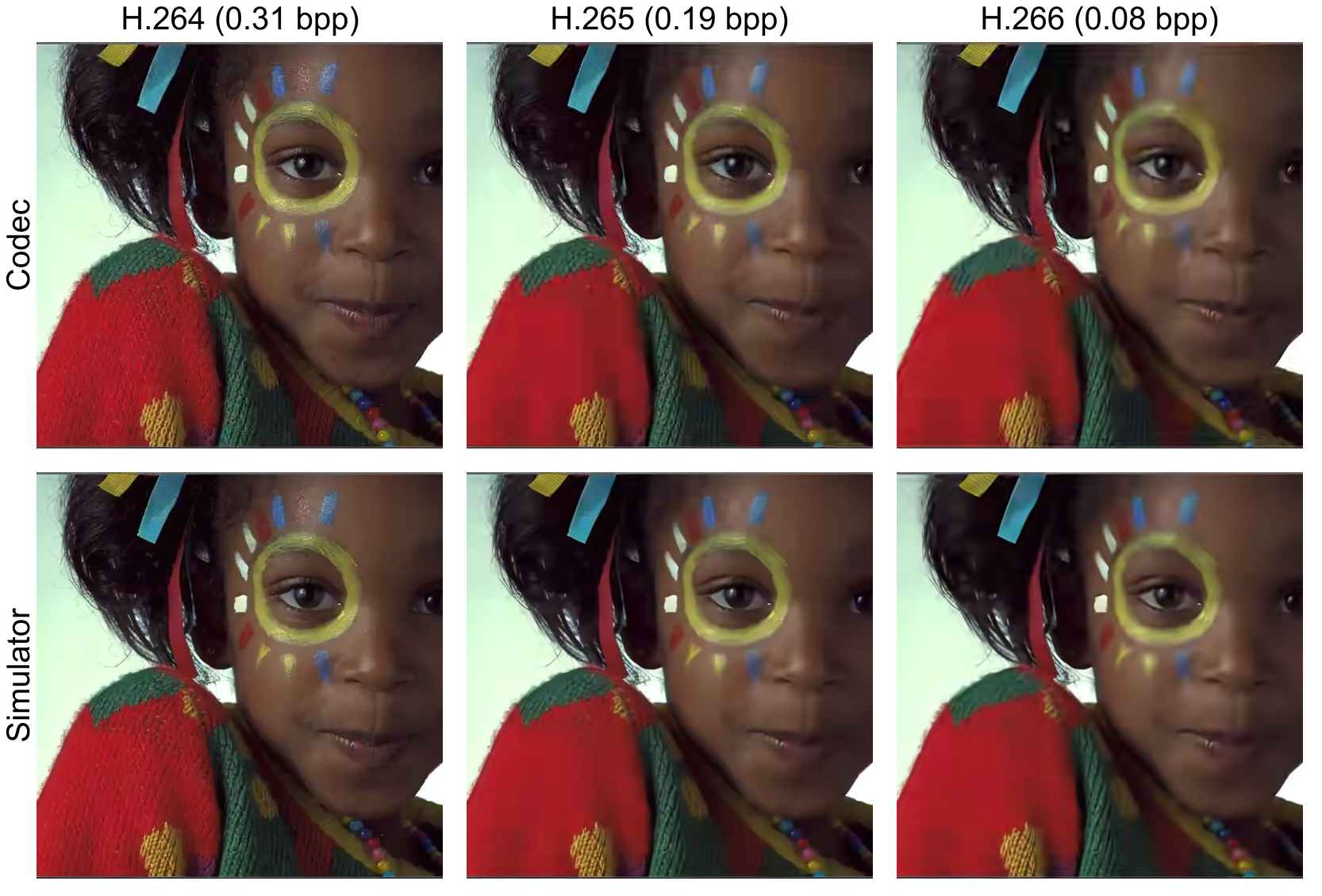}
    \caption{Qualitative results of the simulator under H.264, H.265, and H.266 compression.}
    \label{fig:sim_qual}
\end{figure}

\subsection{Ablation Study on VRPSR Simulator} 
\label{sec:abl}
\textbf{Effectiveness of Simulator Component} In Table~\ref{tab:simu}, we validate the effectiveness of various design components in our simulator. It is shown that using bitrate as input instead of output improves the performance of final SR (Table~\ref{tab:simu} (b) vs (f)). Besides, loading a pre-trained text-to-image diffusion model is also vital to performance (Table~\ref{tab:simu} (a) vs (f)). Further, including codec information as a prompt and embedding is beneficial (Table~\ref{tab:simu} (c) (d) vs (f)). And finally, using a perceptual target instead of an MSE target to train the simulator also improves the perceptual metric of the final SR (Table~\ref{tab:simu} (e) vs (f)).

\textbf{Comparison to Other Simulators} Furthermore, Table~\ref{tab:simu2} shows that our simulator outperforms both JPEG-based and hyperprior-based simulators in terms of perceptual quality. As illustrated in Figure~\ref{fig:sim_qual} and in the supplementary visualization in Figure~\ref{fig:simu2_qual}, our simulator not only produces post-codec results with the best perceptual quality, but its simulated outputs are also the closest to the real codec outputs. This supports the intuition that a more accurate codec simulator leads to better recompression-aware SR models.

\textbf{Relationship Between Simulator Performance and SR Performance} Observing Table~\ref{tab:simu}, an interesting trend is that the PSNR of the simulator does not have much impact on the PSNR of the final SR. However, the perceptual metrics (LPIPS, DISTS, and FID) of the simulator seem to be more related to the perceptual metrics of SR. In Table~\ref{tab:cor}, we quantify this observation by showing the $r^2$ correlation of different metrics between the simulator and SR. In fact, the PSNR of the simulator and SR is only weakly correlated ($r^2=0.105$). However, for LPIPS, DISTS, and FID, the correlation is much stronger ($r^2\ge 0.4$). This coincides with the early works that use differentiable proxy JPEG for training and other codecs for inference \cite{Hu2024StandardCV}. As the target is PSNR/MSE, the accuracy of the simulator is not important. However, for the perceptual target, the accuracy of the simulator becomes important. And as shown in Figure~\ref{fig:sim_qual}, the VRPSR simulator is quite accurate and captures the characteristics of a compressed image.

\begin{table}[thb]
\caption{Correlation between the performance of the simulator and final SR results. Despite the correlation between PSNR of the simulator and SR being weak, the correlation between perceptual metrics of the simulator and SR is strong.}
\label{tab:cor}
\centering
\resizebox{0.5\linewidth}{!}{
\begin{tabular}{@{}ccccc@{}}
\toprule
 & PSNR & LPIPS & DISTS & FID \\ \midrule
$r^2$ btw. Simulator and SR & 0.105 & 0.690 & 0.490 & 0.435 \\ \bottomrule
\end{tabular}
}
\end{table}

\begin{table*}[thb]
\caption{Ablation of training strategies (S3Diff + H.264). Results are reported on the Kodak set.}
\label{tab:abltr}
\centering
\resizebox{\linewidth}{!}{
\begin{tabular}{@{}ccccccccccccc@{}}
\toprule
\multirow{2}{*}{ID} & \multirow{2}{*}{STE} & \multirow{2}{*}{2 Stage Optimization} & \multirow{2}{*}{SR Embed} & \multirow{2}{*}{Versatile Model} & \multirow{2}{*}{Slightly Compressed Supervision} & \multicolumn{5}{c}{S3Diff + H.264} \\
\cmidrule(l){7-11}
& (Sec 3.3.4) & (Sec 3.3.3) & (Sec 3.2.3) & (Sec 3.2.2) & (Sec 3.3.2) & bpp & PSNR & LPIPS & DISTS & FID \\
\midrule
(a) & \cmark & \cmark & \cmark & \cmark & Simulator    & 0.13 & 22.50          & 0.446 & 0.255 & 144.05 \\
(b) & \xmark & \xmark & \cmark & \cmark & Simulator    & 0.13 & 25.03          & 0.419 & 0.257 & 137.12 \\
(c) & \xmark & \cmark & \xmark & \cmark & Simulator    & 0.13 & \textbf{25.74} & 0.412 & 0.237 & 97.06 \\
(d) & \xmark & \cmark & \cmark & \xmark & Simulator    & 0.13 & 25.56          & 0.405 & 0.235 & 92.38 \\
(e) & \xmark & \cmark & \cmark & \cmark & \xmark & 0.13 & 25.18          & 0.427 & 0.242 & 107.82 \\
(f) & \xmark & \cmark & \cmark & \cmark & Real Codec  & 0.13 & 25.68          & 0.415 & 0.240 & 91.02 \\
(g) & \xmark & \cmark & \cmark & \cmark & Simulator    & 0.13 & 25.66          & \textbf{0.404} & \textbf{0.232} & \textbf{89.52} \\
\bottomrule
\end{tabular}
}
\end{table*}

\subsection{Ablation Study on VRPSR Training Technique}

In Table~\ref{tab:abltr}, we show the effectiveness of our proposed training techniques tailored for perceptual SR. More specifically, we show that the additional conditional embedding is beneficial to final performance (Table~\ref{tab:abltr} (c) vs (g)). Besides, we show that adopting the slightly compressed supervision also improves the performance, and using a simulator to generate the compressed supervision is better than using the real codec (Table~\ref{tab:abltr} (e)(f) vs (g)). On the other hand, using STE significantly harms the performance (Table~\ref{tab:abltr} (a) vs (g)). Besides, using two-stage optimization and training the model on versatile conditions helps the performance (Table~\ref{tab:abltr} (b)(d) vs (g)).

\subsection{Generalization to Unseen Codec Conditions}
In the main experiments, VRPSR is trained using codecs operating in constant-quantization-parameter (CQP) mode. In practical deployments, however, modern codecs often adopt dynamic rate-control strategies such as CRF, and may also include codecs that are not seen during training. Therefore, we evaluate whether VRPSR generalizes to (i) an unseen rate-control mode (CRF) and (ii) an unseen codec (AV1), \textbf{without retraining}.

\begin{table}[thb]
\caption{S3Diff with and without VRPSR under x264 CRF rate-control on the Kodak dataset.}
\label{tab:crf_kodak}
\centering
\resizebox{0.6\linewidth}{!}{
\begin{tabular}{@{}lcccccc@{}}
\toprule
Method & CRF & bpp & PSNR & LPIPS & DISTS & FID \\ \midrule
S3Diff            & 36 & 0.10 & 25.67 & 0.477 & 0.280 & 147.46 \\
S3Diff + VRPSR    & 36 & 0.10 & 25.23 & 0.439 & 0.253 & 113.60 \\
S3Diff            & 32 & 0.15 & 26.26 & 0.412 & 0.244 & 103.38 \\
S3Diff + VRPSR    & 32 & 0.15 & 25.94 & 0.382 & 0.220 & 77.81 \\
S3Diff            & 28 & 0.24 & 26.58 & 0.351 & 0.211 & 73.89 \\
S3Diff + VRPSR    & 28 & 0.24 & 26.46 & 0.331 & 0.190 & 62.61 \\
S3Diff            & 26 & 0.35 & 26.69 & 0.322 & 0.193 & 59.08 \\
S3Diff + VRPSR    & 26 & 0.35 & 26.60 & 0.303 & 0.175 & 53.25 \\
\cmidrule(lr){1-7}
& BDBR & -- & 19.36\% & -18.41\% & -27.61\% & -30.10\% \\
\bottomrule
\end{tabular}
}
\end{table}



\textbf{Generalization to Unseen Rate Control (CRF).} We first test VRPSR under the CRF rate-control mode of H.264 (x264), which has never appeared during training. Specifically, we evaluate four CRF values (26, 28, 32, 36), which correspond to average bitrates of approximately $0.10$--$0.35$ bpp on the Kodak dataset. During inference, we measure the resulting bitrate for each image and feed this measured bpp into VRPSR as the rate condition, without any retraining or modification of the model. Table~\ref{tab:crf_kodak} shows consistent improvements on perceptual metrics (LPIPS/DISTS/FID) over the fine-tuned baseline at matched bitrates. This demonstrates that the proposed rate-conditioned simulator enables robust generalization across different rate-control strategies.

\textbf{Generalization to Unseen Codecs.} We further evaluate VRPSR on an unseen codec, AV1, without any retraining. During inference, we reuse the same prompt configuration as H.266 due to similar rate--distortion behavior. As shown in Table~\ref{tab:av1}, VRPSR consistently improves perceptual quality across multiple bitrate settings on both Kodak and ImageNet validation datasets. These results indicate that conditioning the simulator on codec type and bitrate allows VRPSR to generalize beyond the specific codecs used during training, enabling robust recompression-aware super-resolution under diverse deployment conditions.

\begin{table}[thb]
\caption{Direct inference on unseen AV1 (no retraining). Results on Kodak and ImageNet validation sets. Negative BD-rate indicates bitrate saving at the same perceptual quality. For AV1, we reuse the same text prompt as H.266 since their RD are similar.}
\label{tab:av1}
\centering
\resizebox{\linewidth}{!}{
\begin{tabular}{@{}lcccccccccc@{}}
\toprule
 & \multicolumn{5}{c}{Kodak} & \multicolumn{5}{c}{ImageNet (val)} \\
\cmidrule(lr){2-6} \cmidrule(lr){7-11}
 & bpp & PSNR & LPIPS & DISTS & FID & bpp & PSNR & LPIPS & DISTS & FID \\
\midrule
\multirow{4}{*}{S3Diff}
 & 0.09 & 26.25 & 0.471 & 0.257 & 138.08
 & 0.10 & 26.84 & 0.370 & 0.199 & 51.00 \\
 & 0.13 & 26.49 & 0.421 & 0.233 & 103.19
 & 0.14 & 27.12 & 0.327 & 0.178 & 41.03 \\
 & 0.18 & 26.61 & 0.383 & 0.217 & 88.63
 & 0.19 & 27.26 & 0.297 & 0.163 & 35.05 \\
 & 0.24 & 26.68 & 0.345 & 0.197 & 71.55
 & 0.25 & 27.34 & 0.271 & 0.150 & 29.91 \\
\midrule
\multirow{4}{*}{\makecell[l]{S3Diff\\+VRPSR (Ours)}}
 & 0.09 & 25.81 & 0.444 & 0.228 & 107.03
 & 0.10 & 26.42 & 0.341 & 0.177 & 43.29 \\
 & 0.13 & 26.27 & 0.396 & 0.208 & 84.19
 & 0.14 & 26.91 & 0.302 & 0.157 & 34.10 \\
 & 0.18 & 26.51 & 0.365 & 0.193 & 71.22
 & 0.19 & 27.17 & 0.274 & 0.144 & 29.58 \\
 & 0.24 & 26.71 & 0.332 & 0.178 & 56.84
 & 0.25 & 27.33 & 0.251 & 0.134 & 25.89 \\
\cmidrule(lr){2-6} \cmidrule(lr){7-11}
 & BDBR & 28.21\% & \textbf{-15.05\%} & \textbf{-34.66\%} & \textbf{-29.28\%}
 & BDBR & 25.39\% & \textbf{-21.15\%} & \textbf{-33.18\%} & \textbf{-27.63\%} \\

\bottomrule
\end{tabular}
}
\end{table}

\section{Conclusion \& Discussion}
We introduced VRPSR, the first recompression-aware perceptual image super-resolution method. Specifically, we construct a codec simulator that generalizes well to various codec conditions, utilizing a powerful pre-trained diffusion model. Besides, we propose a series of training techniques tailored for perceptual super-resolution, including using a slightly compressed image as supervision and using perceptual targets to train the simulator. Across Real-ESRGAN and S3Diff under H.264/H.265/H.266 single-picture (intra) compression, VRPSR consistently improves perceptual quality at matched bitrates. Beyond preprocessing, our accurate codec simulator also enables joint training of an optional post-processing module.

In this work, we primarily condition the codec with the codec type and target bpp. We further demonstrate that VRPSR generalizes to richer practical codec settings without retraining, including different rate-control modes (e.g., CRF), and unseen codecs (e.g., AV1). Beyond these settings, an interesting direction is to extend VRPSR to even finer-grained codec knobs and to construct a pure codec with clean input, which we leave for future work.

\clearpage  


%
%
\bibliographystyle{splncs04}
\bibliography{main}

\clearpage

\appendix

\section{Implementation Details}
\label{app:impl}

\subsection{Training Details}

\textbf{Hardware \& Software.}
All experiments run on \textbf{8\,$\times$\,NVIDIA A800‑80GB GPUs} (\texttt{CUDA 12.2}), \texttt{Python 3.8.10}, \texttt{Diffusers 0.25.1}, and \path{Accelerate 1.0.1}.

\textbf{Batching Strategy.}
We set the effective global batch size to 64 by adjusting the per-GPU micro-batch size and the number of gradient accumulation steps accordingly.  
For example, a micro-batch size of 1 requires 8 accumulation steps per GPU, while a micro-batch size of 2 requires 4, and so on:
\[
8\;(\text{GPUs}) \times \text{micro‑batch} \times \text{accumulation} = 64.
\]

\textbf{Training Schedules.}
\begin{itemize}
    \item \textbf{Simulator pre‑training:} 60\,000 global steps; 10\,000 global steps (ablation).
    \item \textbf{S3Diff fine‑tuning:} 20\,000 global steps; 10\,000 global steps (ablation).
    \item \textbf{Real-ESRGAN fine‑tuning:} 40\,000 global steps.
\end{itemize}

The learning rate during VRPSR fine-tuning is set to $2\times10^{-5}$ for both Real-ESRGAN and S3Diff. For Real-ESRGAN, we follow the official setting ($1\times10^{-4}$) during its pre-training phase before switching to the VRPSR regime.

\textbf{Optimizer.}
We adopt AdamW with PyTorch defaults: $\beta_1\!=\!0.9$, $\beta_2\!=\!0.999$, $\text{weight\_decay}\!=\!0.01$, and $\varepsilon\!=\!10^{-8}$.

\textbf{Learning‑Rate Scheduler.}
We use the scheduler provided by \path{diffusers.get_scheduler}
(\emph{linear} warm-up $\rightarrow$ \emph{cosine} decay).
We set \texttt{warmup\_steps} and \texttt{max\_train\_steps} to the exact global‑step counts listed above for each stage, ensuring one complete cosine cycle per run.

\textbf{Loss Weights.}
We follow S3Diff's convention and set
$\lambda_{\text{L2}}=2.0$, $\lambda_{\text{LPIPS}}=5.0$, $\lambda_{\text{GAN}}=0.5$
throughout all VRPSR stages.


\subsection{Traditional Codec Details}\label{app:codec}

We compile \texttt{x264} (H.264), \texttt{x265} (H.265), and \texttt{vvenc} (VVC H.266) as shared libraries (\texttt{.so}) and invoke their C APIs from Python.
Each RGB image is converted to YUV 4:2:0, encoded and decoded by the same library, and the decoded reconstruction is used as the final codec output.

\begin{itemize}
    \item \textbf{H.264 / x264}  
          Based on the official x264\footnote{\url{https://code.videolan.org/videolan/x264/}} encoder in
          constant-quantization (CQP) mode, using fixed QP values to control the bitrate.
    \item \textbf{H.265 / x265}  
          Based on the official x265\footnote{\url{https://bitbucket.org/multicoreware/x265_git/src/master/}} implementation with similar settings.
    \item \textbf{H.266 / VVC (vvenc)}  
          Using the vvenc encoder\footnote{\url{https://github.com/fraunhoferhhi/vvenc}}.
\end{itemize}

All three codecs operate in single-picture mode (no B frames) so that each encoded image can be evaluated independently.


\section{Network Architecture}
\label{app:arch}

\begin{figure*}[t]
  \centering
  \includegraphics[width=\linewidth]{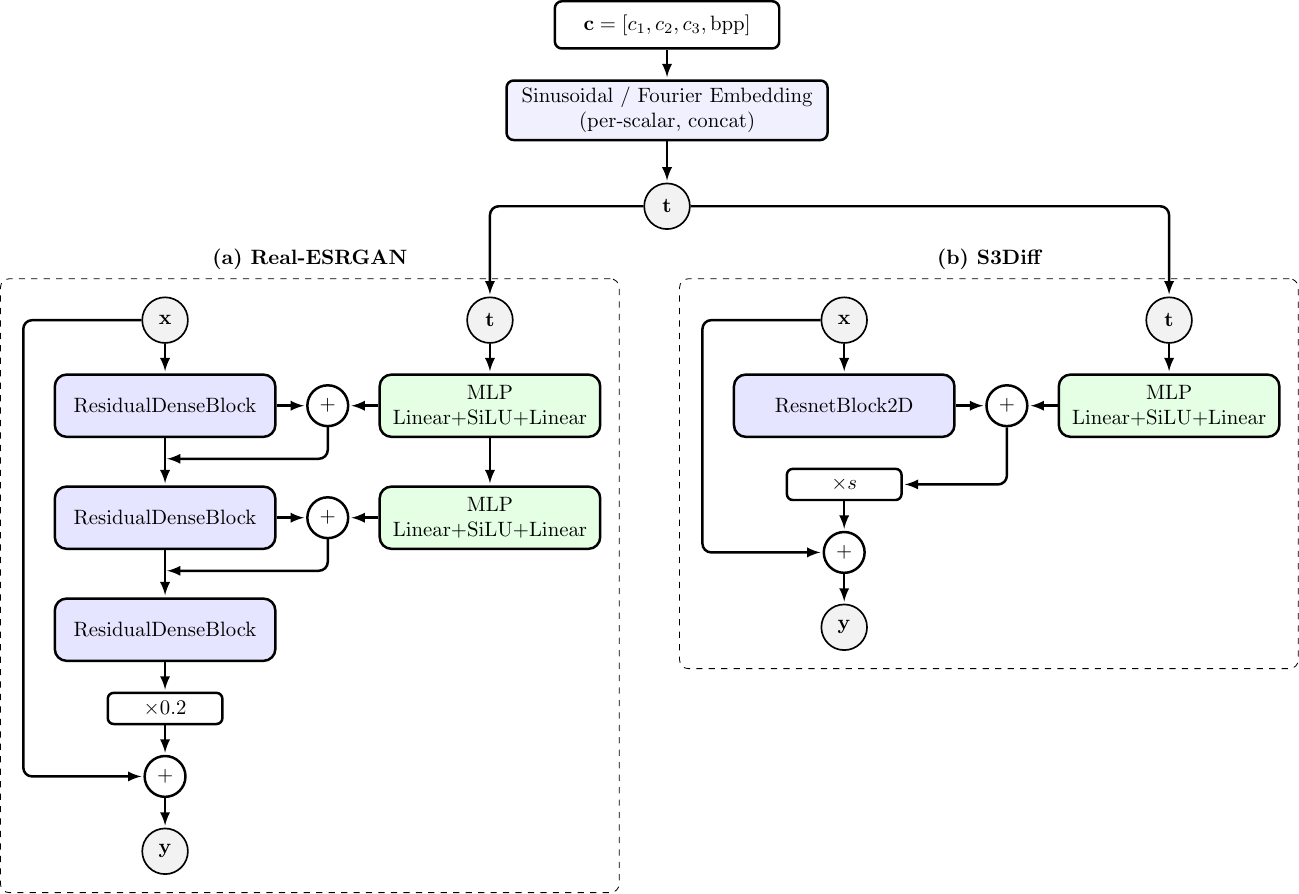}
  \caption{Unified codec-aware conditioning for Real-ESRGAN (left) and S3Diff (right). The 4-D prompt $c=[c_1,c_2,c_3,\mathrm{bpp}]$ is Fourier-embedded per scalar, concatenated, and then injected into the SR backbones as shown.}
  \label{fig:backbones}
\end{figure*}

\subsection{Synthetic Degradation}

We follow Real-ESRGAN to synthesize LR–HR pairs using a \emph{high‑order} degradation process.

First order applies a random degradation chain: $\text{blur} \rightarrow \text{resize} \rightarrow \text{noise} \rightarrow \text{JPEG}$, while the second order repeats the chain with different hyperparameters and applies a sinc filter either before or after JPEG compression to simulate sharpening and expand the degradation diversity.

Such a design has been shown to better match real‑world degradations than a single stage.


\begin{table*}[thb]
\caption{Quantitative results on the ImageNet validation dataset.}
\label{tab:quant1}
\centering
\resizebox{\linewidth}{!}{
\begin{tabular}{@{}lccccccccccccccc@{}}
\toprule
 & \multicolumn{5}{c}{H.264 (x264)} & \multicolumn{5}{c}{H.265 (x265)} & \multicolumn{5}{c}{H.266 (vvenc)} \\ \cmidrule(lr){2-6} \cmidrule(lr){7-11} \cmidrule(lr){12-16}
 & bpp & PSNR & LPIPS & DISTS & FID & bpp & PSNR & LPIPS & DISTS & FID & bpp & PSNR & LPIPS & DISTS & FID \\ \midrule
\multirow{4}{*}{Real-ESRGAN}
 & 0.12 & 26.63 & 0.361 & 0.225 & 56.95 & 0.11 & 27.54 & 0.359 & 0.190 & 51.37 & 0.11 & 28.39 & 0.310 & 0.162 & 36.21 \\
 & 0.18 & 27.60 & 0.302 & 0.189 & 41.51 & 0.15 & 28.27 & 0.309 & 0.162 & 38.14 & 0.16 & 28.64 & 0.290 & 0.151 & 30.58 \\
 & 0.26 & 28.18 & 0.266 & 0.163 & 32.64 & 0.22 & 28.65 & 0.278 & 0.146 & 30.53 & 0.22 & 28.81 & 0.274 & 0.141 & 26.28 \\
 & 0.45 & 28.78 & 0.226 & 0.134 & 23.33 & 0.33 & 28.95 & 0.249 & 0.132 & 24.16 & 0.35 & 29.01 & 0.248 & 0.128 & 21.32 \\
 \midrule
\multirow{5}{*}{\makecell[l]{Real-ESRGAN\\+VRPSR (Ours)}} 
 & 0.12 & 26.32 & 0.345 & 0.210 & 50.08 & 0.11 & 27.02 & 0.346 & 0.179 & 45.95 & 0.11 & 28.11 & 0.296 & 0.153 & 31.46 \\
 & 0.18 & 27.28 & 0.288 & 0.175 & 35.86 & 0.15 & 27.85 & 0.296 & 0.155 & 34.03 & 0.16 & 28.39 & 0.275 & 0.143 & 27.04 \\
 & 0.26 & 27.89 & 0.252 & 0.152 & 27.80 & 0.22 & 28.31 & 0.265 & 0.141 & 27.37 & 0.22 & 28.58 & 0.258 & 0.135 & 23.72 \\
 & 0.45 & 28.50 & 0.211 & 0.128 & 20.75 & 0.33 & 28.64 & 0.234 & 0.128 & 22.25 & 0.35 & 28.75 & 0.228 & 0.123 & 20.05 \\
\cmidrule(lr){2-6} \cmidrule(lr){7-11} \cmidrule(lr){12-16}
& BDBR & 20.83\% & -12.60\% & -16.17\% & -18.07\% & BDBR & 14.88\% & -10.30\% & -10.82\% & -12.53\% & BDBR & 39.16\% & -19.81\% & -17.52\% & -17.44\% \\
\midrule
\multirow{4}{*}{S3Diff}  
 & 0.10 & 25.84 & 0.400 & 0.244 & 65.00 & 0.14 & 26.60 & 0.388 & 0.207 & 57.50 & 0.11 & 27.17 & 0.342 & 0.179 & 43.46 \\
 & 0.14 & 26.63 & 0.337 & 0.205 & 46.69 & 0.20 & 27.12 & 0.331 & 0.176 & 42.07 & 0.16 & 27.38 & 0.304 & 0.159 & 34.33 \\
 & 0.18 & 26.92 & 0.308 & 0.186 & 39.30 & 0.23 & 27.28 & 0.305 & 0.162 & 36.27 & 0.21 & 27.48 & 0.280 & 0.146 & 29.13 \\
 & 0.29 & 27.29 & 0.256 & 0.153 & 27.87 & 0.35 & 27.46 & 0.261 & 0.140 & 26.97 & 0.26 & 27.52 & 0.261 & 0.136 & 24.93 \\
 \midrule
\multirow{5}{*}{\makecell[l]{S3Diff\\+VRPSR (Ours)}} 
 & 0.10 & 25.40 & 0.365 & 0.218 & 52.73 & 0.14 & 26.04 & 0.364 & 0.186 & 48.54 & 0.11 & 26.79 & 0.320 & 0.157 & 34.21 \\
 & 0.14 & 26.26 & 0.306 & 0.180 & 36.55 & 0.20 & 26.76 & 0.310 & 0.157 & 34.05 & 0.16 & 27.16 & 0.287 & 0.142 & 27.95 \\
 & 0.18 & 26.60 & 0.280 & 0.162 & 30.91 & 0.23 & 27.02 & 0.287 & 0.146 & 29.42 & 0.21 & 27.32 & 0.265 & 0.132 & 24.58 \\
 & 0.29 & 27.07 & 0.236 & 0.135 & 23.23 & 0.35 & 27.35 & 0.248 & 0.128 & 23.18 & 0.26 & 27.41 & 0.246 & 0.125 & 21.89 \\
 \cmidrule(lr){2-6} \cmidrule(lr){7-11} \cmidrule(lr){12-16}
& BDBR & 24.01\% & -20.13\% & -25.07\% & -24.48\% & BDBR & 27.23\% & -12.53\% & -20.67\% & -19.49\% & BDBR & 42.02\% & -15.84\% & -30.08\% & -27.45\% \\
\bottomrule
\end{tabular}
}
\end{table*}

\subsection{Differentiable Codec Simulator}

Our simulator is built on top of the official S3Diff codebase\footnote{\url{https://github.com/ArcticHare105/S3Diff}} and reuses the codec-aware conditioning as the S3Diff backbone in Figure~\ref{fig:backbones}. Concretely, the UNet and text encoder remain unchanged; we only add the 4-D codec prompt $c = [c_1,c_2,c_3,\mathrm{bpp}]$ (Sec.~\ref{sec:codec-embed}), which is Fourier-embedded into a vector $\mathbf{t}$ and injected via FiLM-style affine parameters $(\gamma,\beta)$ in each \texttt{ResnetBlock2D}, exactly as in the SR backbone.

Given a super‑resolution image~$X'$ and prompt~$c$, the simulator outputs a rate‑matched reconstruction $\hat{X}'$ that mimics the image produced by the real codec at bitrate~$\bar{R}$. Training follows the same overall objective as the downstream SR model, i.e., a weighted combination of L2, LPIPS, and GAN losses.


\subsection{Super-Resolution Backbones}

\textbf{Codec-aware conditioning (shared).}
Both backbones are made codec-aware by the same 4-D prompt $c=[c_1,c_2,c_3,\mathrm{bpp}]$ (Sec.~\ref{sec:codec-embed}). As shown at the top of Figure~\ref{fig:backbones}, each scalar is first mapped to sinusoidal/Fourier features and then concatenated into a single vector $\mathbf{t}$. This shared embedding $\mathbf{t}$ is then fed into lightweight MLP heads attached to the SR backbones: for Real-ESRGAN, they produce per-channel bias terms that are additively injected into RRDB blocks, while for S3Diff, these heads produce FiLM-style affine parameters $(\gamma,\beta)$. All new MLP layers are zero-initialized in their last linear layer so that, at initialization, the conditioned model exactly matches the pretrained backbone.

\textbf{Real-ESRGAN backbone.}
We start from the official Real-ESRGAN backbone and attach two tiny MLP heads per net that map $\mathbf{t}$ to channel-wise biases, which are added after the first two \texttt{ResidualDenseBlock}s. The third block and the outer residual path remain unchanged.

\textbf{S3Diff backbone.}
We use the original S3Diff UNet and upgrade every \texttt{ResnetBlock2D} to a meta-conditioned block that applies codec-aware FiLM parameters $(\gamma,\beta)$ derived from the prompt $c$ (Sec.~\ref{sec:codec-embed}). No other architectural or training changes are introduced.


\begin{figure*}[thb]
    \centering
    \includegraphics[width=\linewidth]{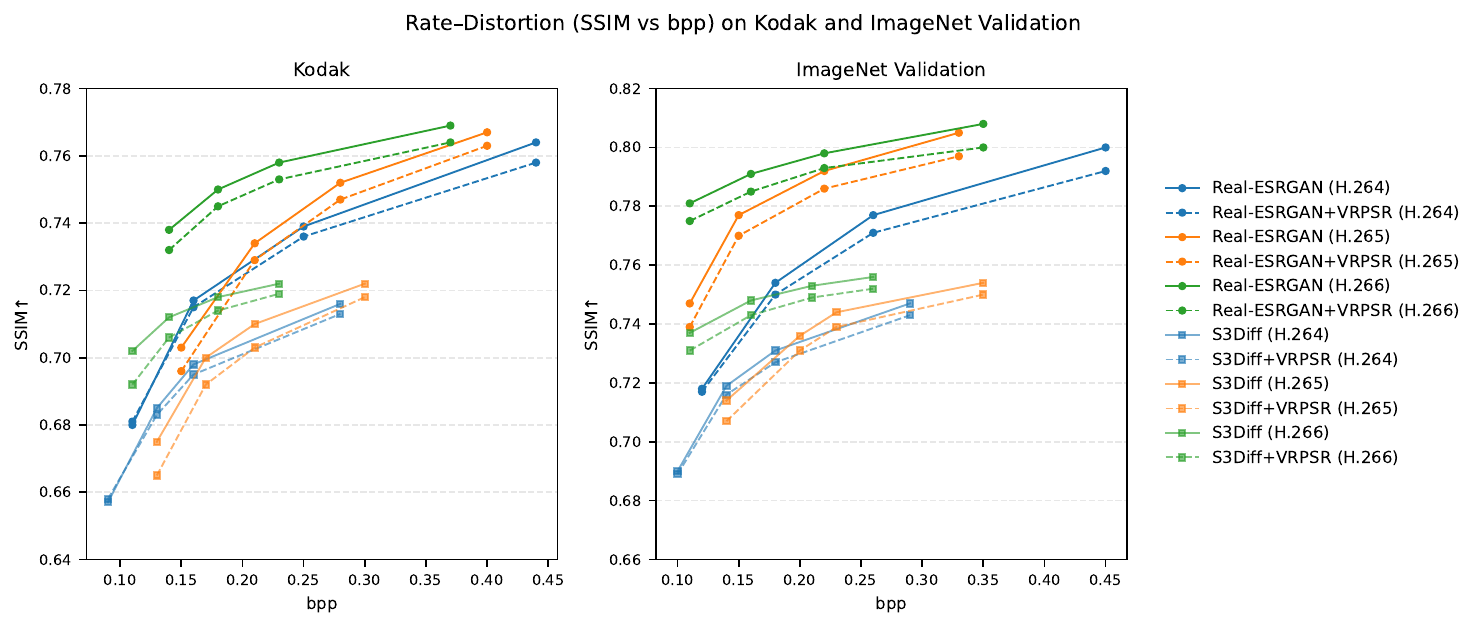}%
    \caption{SSIM-based rate–distortion performance on the Kodak and ImageNet Validation datasets. As a complementary full-reference metric to the perceptual curves in Figure~\ref{fig:rd_curves}, SSIM shows that VRPSR maintains competitive distortion while improving perceptual quality at matched bitrates.}
    \label{fig:rd_curves_ssim}
\end{figure*}

\begin{figure*}[thb]
    \centering
    \includegraphics[width=\linewidth]{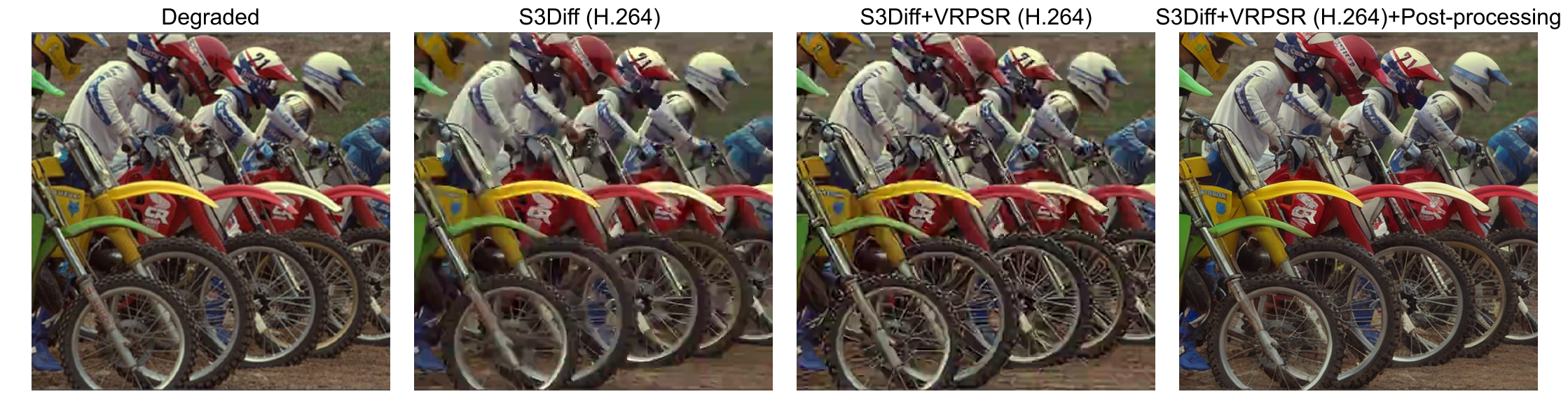}
    \caption{Failure case of VRPSR under extremely heavy compression and its mitigation by the optional post-processing module.}
    \label{fig:fail}
\end{figure*}

\section{Additional Experiments and Analysis}
\label{app:quant}

\subsection{Additional Quantitative Results}
Table~\ref{tab:quant1} reports additional quantitative results on the ImageNet validation split, complementing the main-paper results on Kodak in Table~\ref{tab:quant2}. In addition, Figure~\ref{fig:rd_curves_ssim} presents SSIM-based rate–distortion curves as a complementary full-reference metric to the perceptual RD results in Figure~\ref{fig:rd_curves}.

\subsection{Failure Case and Post-processing}
Under extremely heavy compression (very low bitrate), even VRPSR alone may fail to produce visually sharp results, and residual codec artifacts remain. However, as illustrated in Figure~\ref{fig:fail}, the optional post-processing module introduced in Sec.~\ref{sec:sandwich} can largely remove these artifacts and recover sharper details, effectively mitigating this extreme failure case.

\begin{figure*}[thb]
    \centering
    \includegraphics[width=\linewidth]{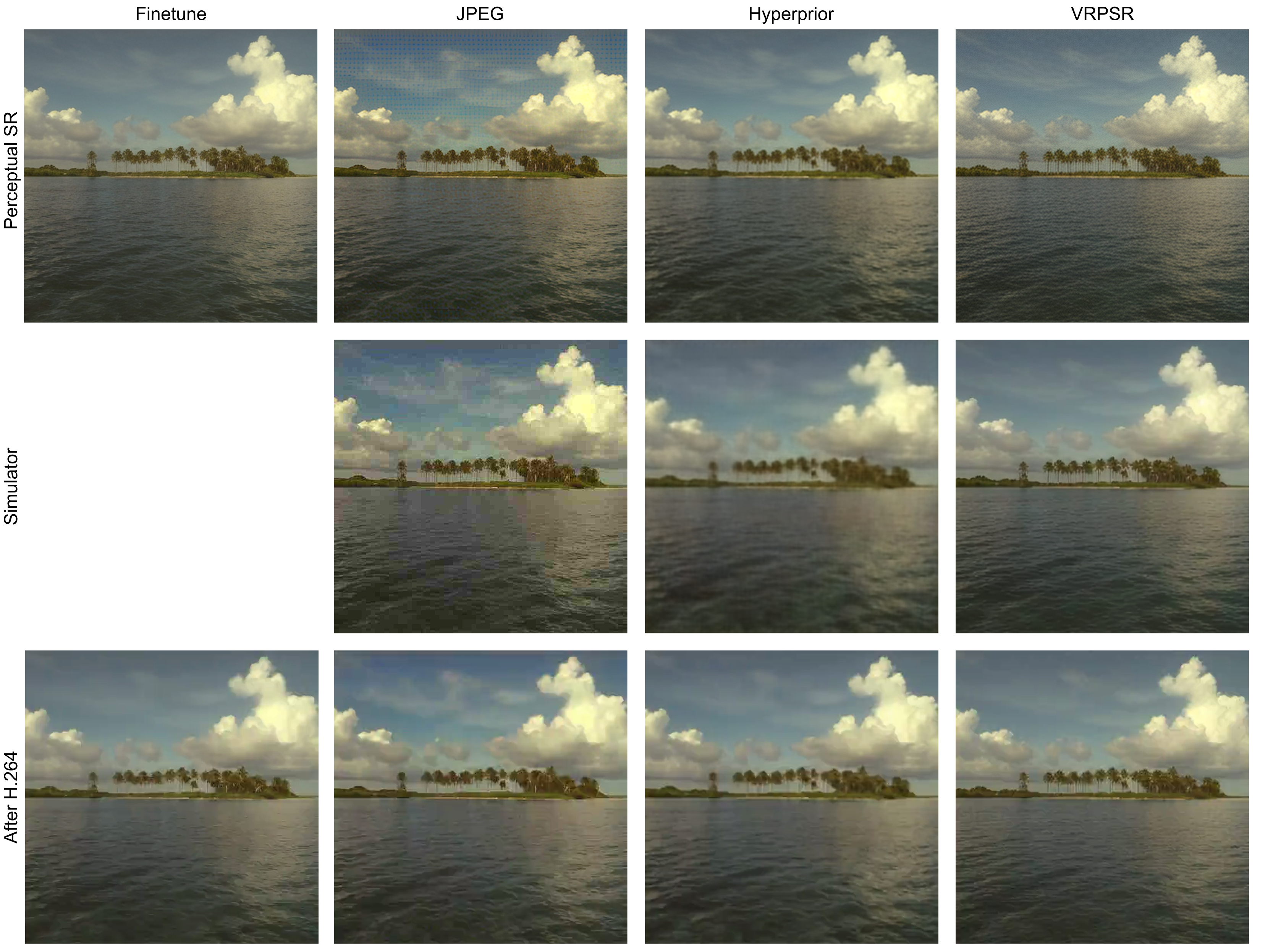}
    \caption{
    Qualitative comparison corresponding to Table~\ref{tab:simu2}.
    Each column shows a super-resolution model trained with a different simulator setting:
    (1) finetuned SR without an explicit codec simulator,
    (2) SR trained with a JPEG-based simulator,
    (3) SR trained with a hyperprior-based simulator,
    and (4) our VRPSR.
    Rows correspond to: 
    (1) the SR output before compression, 
    (2) the output after passing through the corresponding simulator,
    and (3) the output after the real codec.
    Our VRPSR achieves the best final post-codec quality, and its simulated outputs most closely match the real codec outputs, supporting the conclusion that a more accurate simulator leads to better SR models.
    }
    \label{fig:simu2_qual}
\end{figure*}

\subsection{Simulator Comparison}
In Figure~\ref{fig:simu2_qual}, we visualize the setups corresponding to Table~\ref{tab:simu2}.


\subsection{Sensitivity to Supervision Compression Strength}
\label{app:q_sensitivity}

\begin{table}[htb]
\caption{Sensitivity to supervision compression strength. We compare GT supervision with supervision images compressed at different QPs. Lower rows refer to lower-quality supervision images. \textbf{We also include the fine-tuned baseline for reference.} Results are reported on Kodak and ImageNet validation sets.}
\label{tab:qsens}
\centering
\resizebox{0.7\linewidth}{!}{
\begin{tabular}{@{}lcccccccccc@{}}
\toprule
\multirow{2}{*}{Supervision} & \multicolumn{5}{c}{Kodak (S3Diff + H.264)} & \multicolumn{5}{c}{ImageNet (val) (S3Diff + H.264)} \\
\cmidrule(lr){2-6} \cmidrule(lr){7-11}
 & bpp & PSNR & LPIPS & DISTS & FID & bpp & PSNR & LPIPS & DISTS & FID \\
\midrule
\textbf{Finetune (baseline)} & \textbf{0.13} & \textbf{26.05} & \textbf{0.436} & \textbf{0.259} & \textbf{118.89}
                             & \textbf{0.14} & \textbf{26.63} & \textbf{0.337} & \textbf{0.205} & \textbf{46.69} \\
\midrule
GT             & 0.13 & 25.18 & 0.427 & 0.242 & 107.82
               & 0.14 & 25.54 & 0.328 & 0.190 & 39.54 \\
$q{-}25$       & 0.13 & 25.32 & 0.417 & 0.237 & 99.56
               & 0.14 & 25.90 & 0.314 & 0.185 & 38.09 \\
$q{-}20$       & 0.13 & 25.49 & 0.421 & 0.240 & 94.71
               & 0.14 & 26.09 & 0.317 & 0.185 & 37.20 \\
$q{-}15$       & 0.13 & 25.60 & 0.417 & 0.237 & 94.33
               & 0.14 & 26.16 & 0.313 & 0.183 & 37.04 \\
\textbf{$q{-}10$ (main exp.)} & \textbf{0.13} & \textbf{25.66} & \textbf{0.404} & \textbf{0.232} & \textbf{89.52}
               & \textbf{0.14} & \textbf{26.26} & \textbf{0.306} & \textbf{0.180} & \textbf{36.55} \\
$q{-}5$        & 0.13 & 25.93 & 0.410 & 0.242 & 98.74
               & 0.14 & 26.50 & 0.312 & 0.188 & 39.74 \\
$q{-}0$        & 0.13 & 26.17 & 0.427 & 0.253 & 119.79
               & 0.14 & 26.72 & 0.326 & 0.199 & 44.76 \\
\bottomrule
\end{tabular}
}
\end{table}

In Sec.~3.3.2, we propose to supervise the recompression-aware SR model using a slightly compressed image instead of the clean ground truth. The main experiments adopt a supervision level of $q{-}10$. Here we further analyze the sensitivity of this choice.

Table~\ref{tab:qsens} compares supervision images generated using different compression strengths. Specifically, we evaluate supervision images compressed with a range of quantization parameters from $q{-}25$ to $q{-}0$, and compare them with the clean ground truth (GT) supervision as well as the fine-tuned baseline without recompression-aware training.

The results show that a wide range of supervision compression levels consistently outperform GT supervision. In particular, the setting $q{-}10$ achieves the best perceptual metrics (LPIPS, DISTS, and FID) across both Kodak and ImageNet validation datasets. These observations indicate that using slightly compressed supervision is beneficial for recompression-aware SR training, and the performance is robust to the exact compression strength.

\subsection{Statistical Analysis on Kodak and ImageNet}
\label{app:stats}

To complement the aggregate rate--distortion results in Tables~\ref{tab:quant2} and~\ref{tab:simu2}, we report per-image statistics and simple significance tests on both the Kodak and ImageNet validation datasets. For each codec and bitrate setting, we compute PSNR, LPIPS, DISTS, and SSIM for every image and summarize the results as \emph{mean} $\pm$ \emph{standard deviation}. We additionally perform two-sided paired $t$-tests between each baseline SR model and its VRPSR-enhanced counterpart under the same codec and bitrate, using the per-image metrics.

Table~\ref{tab:stats_kodak} reports per-image statistics on Kodak (24 images) for Real-ESRGAN and S3Diff under H.264, H.265, and H.266. Table~\ref{tab:stats_imagenet} shows the corresponding results on the ImageNet validation subset (1{,}000 images). Across almost all codec/bitrate configurations, VRPSR consistently improves the perceptual metrics (LPIPS, DISTS) at comparable bitrates, while PSNR and SSIM changes remain small. This behavior aligns with the perception--distortion trade-off: VRPSR shifts the operating point toward better perceptual quality with only minor fidelity changes. The paired $t$-tests yield extremely small $p$-values for LPIPS and DISTS in nearly all settings (frequently well below $10^{-3}$, and in some ImageNet configurations numerically approaching the machine-precision floor, e.g., $< 10^{-300}$), indicating that the observed perceptual gains are statistically significant rather than arising from random variation. In contrast, the $p$-values for PSNR and SSIM are frequently much larger, confirming that the small average differences in these distortion-oriented metrics are less systematic. The trends on ImageNet largely mirror those on Kodak, suggesting that VRPSR generalizes well from small canonical test sets to more diverse, large-scale natural image distributions.

\section{Reproducibility Statement}
For the experiments, we used publicly accessible datasets. Additional implementation details are provided in the appendix. Moreover, we include the source code for reproducing the experimental results in the supplementary material and will make it publicly available.

\begin{table}[htb]
\caption{Per-image statistics on the Kodak dataset under H.264, H.265, and H.266.}
\label{tab:stats_kodak}
\centering
\resizebox{\linewidth}{!}{
\begin{tabular}{@{}llccccccccccccc@{}}
\toprule
\multirow{3}{*}{Method} & \multirow{3}{*}{Codec} & \multirow{3}{*}{bpp} &
\multicolumn{4}{c}{Finetune SR} &
\multicolumn{4}{c}{VRPSR (Ours)} &
\multicolumn{4}{c}{$p$-value (paired $t$-test)} \\
\cmidrule(lr){4-7} \cmidrule(lr){8-11} \cmidrule(lr){12-15}
 &  &  & PSNR & LPIPS & DISTS & SSIM
 & PSNR & LPIPS & DISTS & SSIM
 & PSNR & LPIPS & DISTS & SSIM \\
\midrule
\multirow{12}{*}{Real-ESRGAN}
 & \multirow{4}{*}{H.264}
   & 0.11 & $26.01 \pm 2.43$ & $0.468 \pm 0.100$ & $0.278 \pm 0.034$ & $0.681 \pm 0.111$
          & $25.80 \pm 2.46$ & $0.451 \pm 0.094$ & $0.264 \pm 0.030$ & $0.681 \pm 0.110$
          & 3.00e-05 & 1.52e-04 & 1.93e-05 & 7.02e-01 \\
 & & 0.16 & $26.86 \pm 2.52$ & $0.403 \pm 0.100$ & $0.243 \pm 0.036$ & $0.717 \pm 0.103$
          & $26.65 \pm 2.50$ & $0.388 \pm 0.095$ & $0.231 \pm 0.032$ & $0.715 \pm 0.102$
          & 1.30e-05 & 2.33e-04 & 1.87e-05 & 1.64e-01 \\
 & & 0.25 & $27.30 \pm 2.48$ & $0.365 \pm 0.095$ & $0.221 \pm 0.035$ & $0.739 \pm 0.095$
          & $27.16 \pm 2.49$ & $0.348 \pm 0.093$ & $0.207 \pm 0.034$ & $0.737 \pm 0.095$
          & 3.86e-05 & 4.33e-06 & 5.24e-09 & 3.11e-02 \\
 & & 0.44 & $27.78 \pm 2.46$ & $0.320 \pm 0.089$ & $0.189 \pm 0.031$ & $0.764 \pm 0.088$
          & $27.64 \pm 2.44$ & $0.296 \pm 0.083$ & $0.180 \pm 0.029$ & $0.759 \pm 0.087$
          & 6.03e-05 & 1.27e-07 & 2.52e-05 & 4.98e-05 \\
\cmidrule(lr){2-15}
 & \multirow{4}{*}{H.265}
   & 0.15 & $26.71 \pm 2.41$ & $0.472 \pm 0.119$ & $0.255 \pm 0.038$ & $0.704 \pm 0.104$
          & $26.23 \pm 2.28$ & $0.456 \pm 0.117$ & $0.238 \pm 0.039$ & $0.696 \pm 0.101$
          & 1.35e-09 & 1.65e-06 & 1.23e-09 & 8.00e-07 \\
 & & 0.21 & $27.32 \pm 2.46$ & $0.419 \pm 0.116$ & $0.225 \pm 0.037$ & $0.734 \pm 0.097$
          & $27.00 \pm 2.37$ & $0.400 \pm 0.116$ & $0.212 \pm 0.037$ & $0.730 \pm 0.095$
          & 1.64e-06 & 5.02e-05 & 3.14e-07 & 6.12e-04 \\
 & & 0.28 & $27.63 \pm 2.45$ & $0.386 \pm 0.111$ & $0.208 \pm 0.036$ & $0.753 \pm 0.092$
          & $27.40 \pm 2.39$ & $0.368 \pm 0.114$ & $0.196 \pm 0.036$ & $0.748 \pm 0.091$
          & 3.43e-05 & 4.26e-06 & 5.86e-08 & 1.90e-06 \\
 & & 0.40 & $27.87 \pm 2.44$ & $0.354 \pm 0.105$ & $0.191 \pm 0.035$ & $0.768 \pm 0.087$
          & $27.71 \pm 2.40$ & $0.328 \pm 0.102$ & $0.179 \pm 0.033$ & $0.763 \pm 0.086$
          & 1.59e-04 & 1.48e-07 & 1.39e-09 & 1.61e-04 \\
\cmidrule(lr){2-15}
 & \multirow{4}{*}{H.266}
   & 0.14 & $27.39 \pm 2.51$ & $0.426 \pm 0.115$ & $0.227 \pm 0.036$ & $0.738 \pm 0.096$
          & $27.14 \pm 2.45$ & $0.409 \pm 0.112$ & $0.214 \pm 0.035$ & $0.733 \pm 0.095$
          & 6.24e-07 & 4.38e-05 & 9.90e-09 & 6.21e-05 \\
 & & 0.18 & $27.60 \pm 2.50$ & $0.403 \pm 0.114$ & $0.215 \pm 0.038$ & $0.751 \pm 0.093$
          & $27.39 \pm 2.46$ & $0.386 \pm 0.112$ & $0.202 \pm 0.037$ & $0.745 \pm 0.093$
          & 7.82e-07 & 9.16e-05 & 7.60e-08 & 4.38e-05 \\
 & & 0.23 & $27.73 \pm 2.51$ & $0.387 \pm 0.113$ & $0.206 \pm 0.039$ & $0.758 \pm 0.091$
          & $27.57 \pm 2.47$ & $0.368 \pm 0.112$ & $0.194 \pm 0.035$ & $0.754 \pm 0.090$
          & 1.80e-05 & 5.84e-06 & 1.34e-06 & 5.68e-05 \\
 & & 0.37 & $27.90 \pm 2.48$ & $0.358 \pm 0.108$ & $0.190 \pm 0.037$ & $0.770 \pm 0.087$
          & $27.78 \pm 2.45$ & $0.328 \pm 0.103$ & $0.178 \pm 0.034$ & $0.764 \pm 0.086$
          & 5.42e-05 & 8.99e-08 & 7.19e-06 & 6.91e-06 \\
\midrule
\multirow{12}{*}{S3Diff}
 & \multirow{4}{*}{H.264}
   & 0.09 & $25.39 \pm 2.45$ & $0.502 \pm 0.106$ & $0.296 \pm 0.031$ & $0.657 \pm 0.122$
          & $24.96 \pm 2.51$ & $0.464 \pm 0.095$ & $0.265 \pm 0.029$ & $0.658 \pm 0.120$
          & 1.51e-08 & 9.36e-07 & 1.59e-11 & 4.48e-01 \\
 & & 0.13 & $26.06 \pm 2.51$ & $0.437 \pm 0.104$ & $0.259 \pm 0.034$ & $0.686 \pm 0.115$
          & $25.67 \pm 2.50$ & $0.404 \pm 0.094$ & $0.232 \pm 0.031$ & $0.683 \pm 0.114$
          & 5.16e-07 & 1.07e-08 & 6.26e-11 & 2.03e-02 \\
 & & 0.16 & $26.31 \pm 2.52$ & $0.403 \pm 0.103$ & $0.241 \pm 0.037$ & $0.699 \pm 0.111$
          & $26.03 \pm 2.55$ & $0.375 \pm 0.095$ & $0.217 \pm 0.033$ & $0.696 \pm 0.113$
          & 1.16e-05 & 4.71e-10 & 9.33e-11 & 7.55e-03 \\
 & & 0.28 & $26.62 \pm 2.51$ & $0.343 \pm 0.096$ & $0.205 \pm 0.035$ & $0.716 \pm 0.106$
          & $26.49 \pm 2.57$ & $0.323 \pm 0.086$ & $0.186 \pm 0.030$ & $0.713 \pm 0.106$
          & 4.35e-03 & 2.36e-06 & 2.62e-09 & 5.54e-03 \\
\cmidrule(lr){2-15}
 & \multirow{4}{*}{H.265}
   & 0.13 & $25.96 \pm 2.36$ & $0.496 \pm 0.116$ & $0.269 \pm 0.036$ & $0.676 \pm 0.111$
          & $25.25 \pm 2.28$ & $0.475 \pm 0.115$ & $0.245 \pm 0.037$ & $0.665 \pm 0.111$
          & 1.82e-11 & 9.33e-05 & 1.72e-10 & 1.44e-06 \\
 & & 0.17 & $26.43 \pm 2.42$ & $0.437 \pm 0.121$ & $0.236 \pm 0.039$ & $0.701 \pm 0.107$
          & $25.99 \pm 2.37$ & $0.414 \pm 0.114$ & $0.215 \pm 0.037$ & $0.693 \pm 0.107$
          & 6.70e-08 & 4.77e-05 & 1.88e-09 & 9.29e-06 \\
 & & 0.21 & $26.59 \pm 2.43$ & $0.409 \pm 0.117$ & $0.222 \pm 0.037$ & $0.710 \pm 0.105$
          & $26.26 \pm 2.41$ & $0.385 \pm 0.111$ & $0.200 \pm 0.036$ & $0.704 \pm 0.105$
          & 1.92e-06 & 1.52e-06 & 3.61e-10 & 1.02e-06 \\
 & & 0.30 & $26.77 \pm 2.44$ & $0.354 \pm 0.106$ & $0.194 \pm 0.035$ & $0.723 \pm 0.101$
          & $26.64 \pm 2.46$ & $0.342 \pm 0.101$ & $0.180 \pm 0.031$ & $0.718 \pm 0.102$
          & 2.88e-03 & 5.32e-03 & 8.25e-08 & 1.91e-04 \\
\cmidrule(lr){2-15}
 & \multirow{4}{*}{H.266}
   & 0.11 & $26.50 \pm 2.49$ & $0.446 \pm 0.118$ & $0.243 \pm 0.034$ & $0.702 \pm 0.108$
          & $25.98 \pm 2.45$ & $0.423 \pm 0.115$ & $0.211 \pm 0.030$ & $0.692 \pm 0.107$
          & 3.55e-06 & 9.40e-04 & 6.34e-11 & 2.82e-04 \\
 & & 0.14 & $26.64 \pm 2.48$ & $0.413 \pm 0.118$ & $0.223 \pm 0.037$ & $0.712 \pm 0.105$
          & $26.37 \pm 2.47$ & $0.390 \pm 0.111$ & $0.197 \pm 0.033$ & $0.706 \pm 0.104$
          & 7.86e-06 & 5.93e-04 & 8.86e-10 & 5.57e-04 \\
 & & 0.18 & $26.73 \pm 2.48$ & $0.385 \pm 0.114$ & $0.209 \pm 0.038$ & $0.718 \pm 0.103$
          & $26.57 \pm 2.49$ & $0.367 \pm 0.108$ & $0.186 \pm 0.035$ & $0.714 \pm 0.102$
          & 3.88e-05 & 2.25e-03 & 1.72e-08 & 2.85e-04 \\
 & & 0.23 & $26.78 \pm 2.47$ & $0.362 \pm 0.115$ & $0.195 \pm 0.039$ & $0.723 \pm 0.102$
          & $26.69 \pm 2.50$ & $0.346 \pm 0.110$ & $0.175 \pm 0.033$ & $0.719 \pm 0.102$
          & 4.01e-03 & 1.77e-03 & 1.37e-08 & 4.60e-04 \\
\bottomrule
\end{tabular}
}
\end{table}

\begin{table}[]
\caption{Per-image statistics on the ImageNet Validation dataset under H.264, H.265, and H.266.}
\label{tab:stats_imagenet}
\centering
\resizebox{\linewidth}{!}{
\begin{tabular}{@{}llccccccccccccc@{}}
\toprule
\multirow{3}{*}{Method} & \multirow{3}{*}{Codec} & \multirow{3}{*}{bpp} &
\multicolumn{4}{c}{Finetune SR} &
\multicolumn{4}{c}{VRPSR (Ours)} &
\multicolumn{4}{c}{$p$-value (paired $t$-test)} \\
\cmidrule(lr){4-7} \cmidrule(lr){8-11} \cmidrule(lr){12-15}
 &  &  & PSNR & LPIPS & DISTS & SSIM
 & PSNR & LPIPS & DISTS & SSIM
 & PSNR & LPIPS & DISTS & SSIM \\
\midrule
\multirow{12}{*}{Real-ESRGAN}
 & \multirow{4}{*}{H.264}
   & 0.12 & $26.64 \pm 3.55$ & $0.362 \pm 0.123$ & $0.225 \pm 0.037$ & $0.719 \pm 0.142$
          & $26.33 \pm 3.46$ & $0.345 \pm 0.123$ & $0.210 \pm 0.035$ & $0.717 \pm 0.143$
          & 1.13e-172 & 1.07e-139 & 1.07e-227 & 1.82e-11 \\
 & & 0.18 & $27.61 \pm 3.76$ & $0.302 \pm 0.120$ & $0.189 \pm 0.039$ & $0.755 \pm 0.132$
          & $27.28 \pm 3.64$ & $0.289 \pm 0.120$ & $0.175 \pm 0.036$ & $0.750 \pm 0.134$
          & 3.81e-209 & 4.54e-141 & 5.32e-196 & 1.20e-82 \\
 & & 0.26 & $28.19 \pm 3.89$ & $0.266 \pm 0.117$ & $0.164 \pm 0.039$ & $0.777 \pm 0.126$
          & $27.90 \pm 3.79$ & $0.253 \pm 0.117$ & $0.153 \pm 0.038$ & $0.771 \pm 0.128$
          & 8.71e-202 & 8.56e-153 & 3.99e-161 & 2.46e-122 \\
 & & 0.45 & $28.79 \pm 4.05$ & $0.226 \pm 0.113$ & $0.135 \pm 0.039$ & $0.800 \pm 0.120$
          & $28.51 \pm 3.97$ & $0.211 \pm 0.110$ & $0.128 \pm 0.039$ & $0.792 \pm 0.122$
          & 3.35e-181 & 1.69e-144 & 1.68e-93 & 3.87e-166 \\
\cmidrule(lr){2-15}
 & \multirow{4}{*}{H.265}
   & 0.11 & $27.55 \pm 3.64$ & $0.359 \pm 0.138$ & $0.191 \pm 0.042$ & $0.748 \pm 0.133$
          & $27.02 \pm 3.38$ & $0.346 \pm 0.144$ & $0.180 \pm 0.041$ & $0.740 \pm 0.134$
          & 2.43e-186 & 8.23e-62 & 1.21e-132 & 7.06e-125 \\
 & & 0.15 & $28.28 \pm 3.84$ & $0.309 \pm 0.134$ & $0.163 \pm 0.041$ & $0.777 \pm 0.126$
          & $27.86 \pm 3.62$ & $0.297 \pm 0.138$ & $0.155 \pm 0.041$ & $0.770 \pm 0.127$
          & 4.76e-151 & 5.84e-76 & 9.63e-89 & 3.33e-139 \\
 & & 0.22 & $28.66 \pm 3.96$ & $0.279 \pm 0.131$ & $0.147 \pm 0.041$ & $0.793 \pm 0.122$
          & $28.31 \pm 3.78$ & $0.266 \pm 0.132$ & $0.141 \pm 0.040$ & $0.786 \pm 0.123$
          & 8.27e-141 & 8.06e-89 & 1.09e-57 & 2.09e-139 \\
 & & 0.33 & $28.95 \pm 4.08$ & $0.250 \pm 0.125$ & $0.132 \pm 0.040$ & $0.805 \pm 0.119$
          & $28.65 \pm 3.97$ & $0.234 \pm 0.124$ & $0.129 \pm 0.040$ & $0.797 \pm 0.121$
          & 2.92e-151 & 3.03e-106 & 1.30e-34 & 2.94e-144 \\
\cmidrule(lr){2-15}
 & \multirow{4}{*}{H.266}
   & 0.11 & $28.40 \pm 3.99$ & $0.311 \pm 0.133$ & $0.162 \pm 0.040$ & $0.781 \pm 0.126$
          & $28.12 \pm 3.89$ & $0.296 \pm 0.133$ & $0.153 \pm 0.039$ & $0.775 \pm 0.128$
          & 4.93e-164 & 1.64e-111 & 3.22e-117 & 5.12e-114 \\
 & & 0.16 & $28.65 \pm 4.05$ & $0.290 \pm 0.129$ & $0.151 \pm 0.040$ & $0.792 \pm 0.123$
          & $28.40 \pm 3.95$ & $0.276 \pm 0.129$ & $0.143 \pm 0.040$ & $0.785 \pm 0.125$
          & 1.06e-145 & 1.89e-133 & 2.98e-120 & 1.68e-127 \\
 & & 0.22 & $28.82 \pm 4.09$ & $0.274 \pm 0.127$ & $0.142 \pm 0.040$ & $0.799 \pm 0.121$
          & $28.58 \pm 3.99$ & $0.259 \pm 0.125$ & $0.136 \pm 0.040$ & $0.793 \pm 0.122$
          & 5.09e-136 & 1.17e-131 & 1.88e-83 & 1.68e-138 \\
 & & 0.35 & $29.01 \pm 4.14$ & $0.249 \pm 0.123$ & $0.129 \pm 0.040$ & $0.808 \pm 0.119$
          & $28.76 \pm 4.06$ & $0.228 \pm 0.117$ & $0.124 \pm 0.040$ & $0.800 \pm 0.120$
          & 4.62e-157 & 2.05e-154 & 1.75e-63 & 1.02e-175 \\
\midrule
\multirow{12}{*}{S3Diff}
 & \multirow{4}{*}{H.264}
   & 0.10 & $25.84 \pm 3.43$ & $0.400 \pm 0.126$ & $0.244 \pm 0.036$ & $0.690 \pm 0.153$
          & $25.41 \pm 3.39$ & $0.365 \pm 0.122$ & $0.218 \pm 0.036$ & $0.689 \pm 0.152$
          & 5.14e-226 & 1.82e-254 & $<$1e-300 & 4.68e-02 \\
 & & 0.14 & $26.63 \pm 3.66$ & $0.337 \pm 0.123$ & $0.206 \pm 0.037$ & $0.720 \pm 0.147$
          & $26.26 \pm 3.61$ & $0.307 \pm 0.117$ & $0.180 \pm 0.034$ & $0.716 \pm 0.148$
          & 5.60e-220 & 8.73e-252 & $<$1e-300 & 3.24e-46 \\
 & & 0.18 & $26.92 \pm 3.76$ & $0.309 \pm 0.120$ & $0.187 \pm 0.038$ & $0.731 \pm 0.146$
          & $26.61 \pm 3.71$ & $0.280 \pm 0.114$ & $0.163 \pm 0.034$ & $0.727 \pm 0.146$
          & 6.60e-180 & 1.43e-243 & 1.48e-309 & 5.70e-64 \\
 & & 0.29 & $27.29 \pm 3.91$ & $0.257 \pm 0.112$ & $0.153 \pm 0.037$ & $0.747 \pm 0.143$
          & $27.08 \pm 3.85$ & $0.236 \pm 0.106$ & $0.135 \pm 0.034$ & $0.743 \pm 0.144$
          & 2.90e-98 & 3.69e-171 & 4.95e-250 & 5.48e-81 \\
\cmidrule(lr){2-15}
 & \multirow{4}{*}{H.265}
   & 0.14 & $26.61 \pm 3.56$ & $0.389 \pm 0.139$ & $0.208 \pm 0.042$ & $0.714 \pm 0.148$
          & $26.04 \pm 3.39$ & $0.364 \pm 0.141$ & $0.187 \pm 0.041$ & $0.707 \pm 0.146$
          & 3.20e-231 & 1.38e-149 & 2.20e-257 & 2.78e-91 \\
 & & 0.20 & $27.12 \pm 3.77$ & $0.331 \pm 0.133$ & $0.176 \pm 0.040$ & $0.737 \pm 0.144$
          & $26.77 \pm 3.62$ & $0.311 \pm 0.132$ & $0.158 \pm 0.038$ & $0.731 \pm 0.143$
          & 3.61e-158 & 5.34e-142 & 2.32e-250 & 8.97e-114 \\
 & & 0.23 & $27.28 \pm 3.84$ & $0.306 \pm 0.129$ & $0.162 \pm 0.039$ & $0.745 \pm 0.143$
          & $27.02 \pm 3.73$ & $0.288 \pm 0.127$ & $0.146 \pm 0.036$ & $0.739 \pm 0.143$
          & 1.51e-110 & 5.22e-125 & 7.66e-218 & 5.23e-112 \\
 & & 0.35 & $27.47 \pm 3.96$ & $0.262 \pm 0.120$ & $0.140 \pm 0.037$ & $0.754 \pm 0.142$
          & $27.35 \pm 3.89$ & $0.248 \pm 0.116$ & $0.129 \pm 0.035$ & $0.750 \pm 0.142$
          & 1.06e-36 & 6.71e-96 & 8.88e-145 & 1.35e-83 \\
\cmidrule(lr){2-15}
 & \multirow{4}{*}{H.266}
   & 0.11 & $27.18 \pm 3.87$ & $0.342 \pm 0.135$ & $0.180 \pm 0.038$ & $0.737 \pm 0.146$
          & $26.79 \pm 3.73$ & $0.320 \pm 0.132$ & $0.157 \pm 0.034$ & $0.731 \pm 0.145$
          & 8.59e-150 & 6.47e-125 & 4.91e-261 & 2.61e-79 \\
 & & 0.16 & $27.39 \pm 3.93$ & $0.305 \pm 0.128$ & $0.160 \pm 0.037$ & $0.748 \pm 0.144$
          & $27.16 \pm 3.84$ & $0.288 \pm 0.125$ & $0.142 \pm 0.034$ & $0.743 \pm 0.143$
          & 8.17e-96 & 7.37e-111 & 2.31e-232 & 1.12e-88 \\
 & & 0.21 & $27.48 \pm 3.97$ & $0.280 \pm 0.122$ & $0.147 \pm 0.036$ & $0.754 \pm 0.143$
          & $27.33 \pm 3.91$ & $0.265 \pm 0.119$ & $0.133 \pm 0.034$ & $0.749 \pm 0.143$
          & 4.76e-62 & 2.01e-107 & 1.08e-199 & 2.13e-83 \\
 & & 0.26 & $27.53 \pm 4.00$ & $0.262 \pm 0.118$ & $0.137 \pm 0.035$ & $0.757 \pm 0.142$
          & $27.41 \pm 3.95$ & $0.247 \pm 0.113$ & $0.126 \pm 0.034$ & $0.752 \pm 0.142$
          & 2.02e-44 & 5.65e-109 & 2.91e-172 & 1.50e-82 \\
\bottomrule
\end{tabular}
}
\end{table}

\end{document}